\documentclass[10pt,twocolumn,letterpaper]{article}

\usepackage{cvpr}              

\definecolor{cvprblue}{rgb}{0.21,0.49,0.74}
\usepackage[pagebackref,breaklinks,colorlinks,allcolors=cvprblue]{hyperref}

\usepackage{mathtools}
\usepackage{xfrac}
\usepackage[utf8]{inputenc} 
\usepackage[T1]{fontenc}    
\usepackage{hyperref}       
\usepackage{url}            
\usepackage{booktabs}       
\usepackage{amsfonts}       
\usepackage{nicefrac}       
\usepackage{microtype}      
\usepackage{xcolor}         
\usepackage{amsmath}
\usepackage{amssymb}
\usepackage{mathtools}
\usepackage{amsthm}
\usepackage{colortbl}
\usepackage{xcolor}
\usepackage{makecell}
\usepackage{lipsum}
\usepackage{algorithm}
\usepackage{algorithmic}
\usepackage{multirow}
\usepackage[misc]{ifsym}

\definecolor{mygreen}{rgb}{0, 0.6823, 0.7215}
\definecolor{GoogleRed}{RGB}{234, 67, 53}
\definecolor{GoogleBlue}{RGB}{66, 133, 244}
\definecolor{GoogleGreen}{RGB}{52, 168, 83}

\definecolor{diamondPurple}{RGB}{103, 58, 183}
\definecolor{myblue}{RGB}{47, 114, 173}
\definecolor{PINK}{RGB}{252, 81, 133}
\hypersetup{
  urlcolor = PINK,
  linkcolor = myblue,
  citecolor  = myblue,
  colorlinks = true,
}
\newcommand{\eps}{\boldsymbol{\epsilon}}
\newcommand{\noise}{\mathcal{N}(\boldsymbol{0},\mathbf{I})}

\def\paperID{10579} 
\def\confName{CVPR}
\def\confYear{2025}

\title{Calibrated Multi-Preference Optimization for Aligning Diffusion Models}

\author{
Kyungmin Lee$^{1, 2, \dagger \textsuperscript{\Letter}}$\quad 
Xiaohang Li$^3$\quad
Qifei Wang$^1$\quad
Junfeng He$^4$\quad
Junjie Ke$^1$ \\
Ming-Hsuan Yang$^{1}$\quad
Irfan Essa$^{1, 5}$\quad
Jinwoo Shin$^2$\quad
Feng Yang$^{1, \ddagger}$\quad
Yinxiao Li$^{1, \ddagger \textsuperscript{\Letter}}$ \\
\\
$^1$Google DeepMind\quad
$^2$KAIST\quad
$^3$Google\quad
$^4$Google Research\quad
$^5$Georgia Institute of Technology
}

\begin{document}
\maketitle

\begingroup
\renewcommand{\thefootnote}{\fnsymbol{footnote}}

\footnotetext[2]{Work done during an internship at Google DeepMind.}
\footnotetext{\hspace{-3.5mm} \textsuperscript{\Letter} Corresponding Authors. \hspace{2mm}$\ddagger$ Equal advising.} 
\endgroup

\begin{abstract}
Aligning text-to-image (T2I) diffusion models with preference optimization is valuable for human-annotated datasets, but the heavy cost of manual data collection limits scalability. Using reward models offers an alternative, however, current preference optimization methods fall short in exploiting the rich information, as they only consider pairwise preference distribution. Furthermore, they lack generalization to multi-preference scenarios and struggle to handle inconsistencies between rewards. To address this, we present Calibrated Preference Optimization (CaPO), a novel method to align T2I diffusion models by incorporating the general preference from multiple reward models without human annotated data. The core of our approach involves a reward calibration method to approximate the general preference by computing the expected win-rate against the samples generated by the pretrained models. Additionally, we propose a frontier-based pair selection method that effectively manages the multi-preference distribution by selecting pairs from Pareto frontiers. Finally, we use regression loss to fine-tune diffusion models to match the difference between calibrated rewards of a selected pair. Experimental results show that CaPO consistently outperforms prior methods, such as Direct Preference Optimization (DPO), in both single and multi-reward settings validated by evaluation on T2I benchmarks, including GenEval and T2I-Compbench.\footnote{\textbf{Project page}: \url{https://kyungmnlee.github.io/capo.github.io/}}
\end{abstract}    
\section{Introduction}\label{sec:intro}
Recent text-to-image (T2I) diffusion models~\citep{saharia2022photorealistic, nichol2021glide, rombach2022high, esser2024scaling, betker2023improving} generate high-quality images from text prompts. While these models perform well, synthesizing images that closely match subtle human preferences is a challenging task. Following the success of reinforcement learning from human feedback (RLHF) in language models~\citep{ouyang2022training}, training a reward model to mimic human preference~\citep{lee2023aligning, kirstain2023pick, xu2023imagereward, wu2023human, wu2023human2, zhang2024learning}, and fine-tuning diffusion models with RL algorithms shows promise~\citep{black2023training, fan2023dpok, deng2024prdp, lee2025parrot}. However, the computational expense of backpropagation through the diffusion trajectories limits the scalability to large-scale diffusion models. 
To address this problem, Diffusion-DPO~\citep{wallace2023diffusion} applies direct preference optimization (DPO)~\citep{rafailov2024direct} to diffusion models, with good results for large-scale diffusion models~\citep{esser2024scaling}. Nonetheless, since Diffusion-DPO entails an expensive paired human preference dataset, it remains unclear how to leverage multiple reward models to align large-scale T2I diffusion models.

Building on this line of research, we explore an alternative approach to fine-tune large-scale T2I diffusion models without relying on human preference datasets. Instead, we generate training data using pretrained T2I diffusion models and simulate human preferences through multiple reward models. Unlike Diffusion-DPO~\citep{wallace2023diffusion}, which relies on explicit pairwise preference data, our method fully leverages the rich knowledge embedded in reward signals. However, directly optimizing rewards risks overfitting and reward hacking if the reward values are not properly calibrated~\citep{gao2023scaling}.

\begin{figure*}[t]
    \small\centering
    \includegraphics[width=0.98\linewidth]{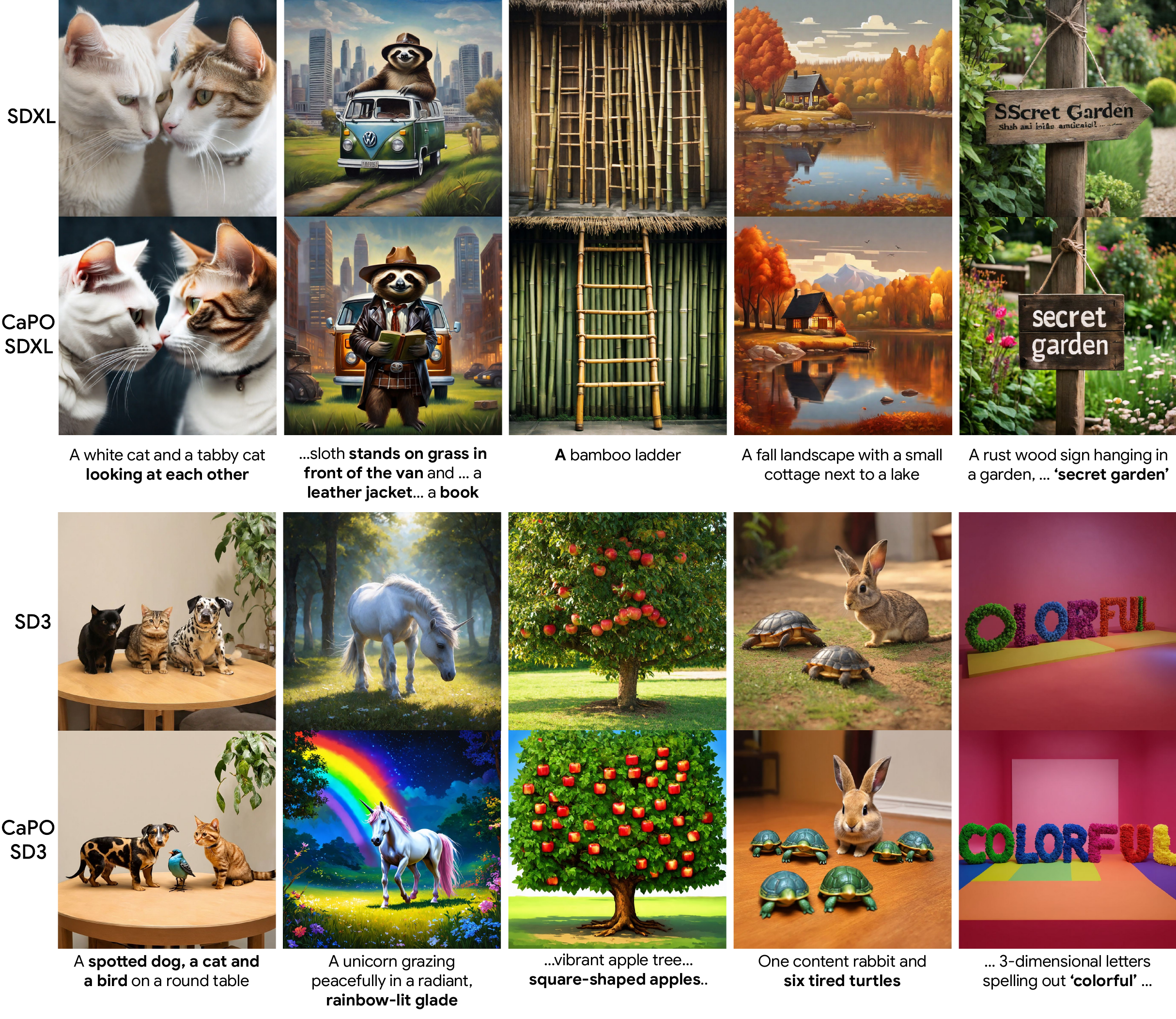}
    \vspace{-0.1in}
    \caption{
    Calibrated Preference Optimization (CaPO) improves the performance of diffusion models by optimizing the model with diverse reward signals. The top and bottom groups are using SDXL and SD3-medium, respectively. For each group, the first row is from base model and the second row is applying CaPO to the base model. CaPO tends to generate images of higher quality (\emph{e.g.}, image aestheticism, text rendering), and better prompt alignment (\emph{e.g.}, compositional generation), without using any human preference dataset.
    }
    \label{fig:teaser}
    \vspace{-0.1in}
\end{figure*}
To address this issue, we propose \emph{Calibrated Preference Optimization} (CaPO) to enhance preference optimization of T2I diffusion models by improving how reward signals are used. 
Instead of directly optimizing reward values, we introduce the concept of general preference~\citep{azar2024general}, defined as the expected win-rate against a pretrained model. We approximate this by averaging pairwise win-rates among multiple samples, providing a robust and calibrated signal. Our fine-tuning objective uses the regression loss to match the difference of calibrated rewards with the difference of implicit reward from diffusion models, which is simple and effective that enhances the performance.
In addition, we introduce a novel Frontier-based rejection sampling method to enhance the multi-reward preference optimization.
This approach addresses the limitations of combining rewards with linear weights~\citep{clark2023directly, deng2024prdp} by selecting training pairs from the upper and lower Pareto frontiers using a non-dominated sorting algorithm. Jointly optimizing diverse reward signals enables the model to achieve balanced response to multiple rewards and mitigate the over-optimization problem when using a single reward. Lastly, we propose an effective loss weighting scheme to improve the diffusion preference optimization.

Through extensive experiments, we show that CaPO consistently outperforms other fine-tuning methods including DPO~\citep{wallace2023diffusion}, achieving better alignment with human preferences across different benchmarks. Our contributions are:
\begin{itemize}
    \item We propose CaPO, which leverages a novel reward calibration method by incorporating approximated win-rates to fine-tune diffusion models and mitigate reward hacking;
    \item We expand the applicability of CaPO to multi-reward fine-tuning problems by introducing frontier-based rejection sampling to jointly optimize with diverse reward signals; 
    \item We demonstrate the effectiveness of CaPO with favorable visual generation quality against state-of-the-art models on benchmark datasets.
\end{itemize}

\begin{figure*}[t]
    \small\centering
    \includegraphics[width=1.0\textwidth]{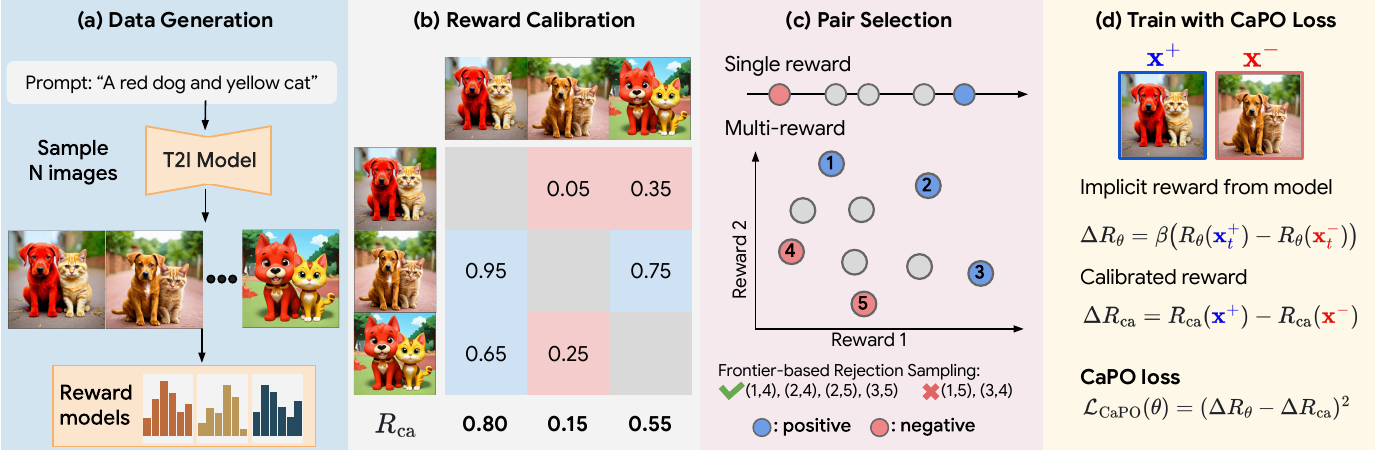}
    \caption{
    \textbf{Overview}. 
    (a) We generate $N$ images using pretrained T2I diffusion model using the prompt dataset, and infer the scores from reward models. (b) Then, we calibrate the rewards by making pairwise comparison between images. For each image, we compute the win-rates between other $N-1$ images using Eq.~\eqref{eq:bt}, and average them to obtain calibrated reward $R_{\textrm{ca}}$ (see Sec.~\ref{sec:capo}). (c) We select pair by choosing the best-of-$N$ and worst-of-$N$ when using single reward. For multi-reward, we use non-dominated sorting algorithm to select upper Pareto set as positives, and lower Pareto set as negatives. The accepted and rejected pairs are also listed using proposed rejection sampling method. (d) Lastly, during training, we select a pair from (c), and compute CaPO loss (\emph{i.e.}, Eq.~\eqref{eq:capoloss}), which perform regression task to match the difference in calibrated rewards, \emph{i.e.}, $\Delta R_{\textrm{ca}}$ by the difference of implicit reward model $\Delta R_\theta$.
    }
    \label{fig:overview}
    \vspace{-10pt}
\end{figure*}

\section{Related Work}
\label{sec:relatedworks}
\vspace{-0.1in}
\noindent
{\bf Modeling human preference for visual generation.}
Motivated by the recent success in incorporating human preference modeling to convert language models into advanced chatbot models, numerous methods have been developed to transfer the success into visual generation. \citet{lee2023aligning} demonstrates the capability of training reward models according to human preference using a small dataset. Subsequently, numerous reward models~\citep{kirstain2023pick, xu2023imagereward, wu2023human, wu2023human2, zhang2024learning} for text-to-image generation have been proposed by fine-tuning a vision-language model (\emph{e.g.}, CLIP~\citep{radford2021learning} or BLIP~\citep{li2022blip}) with Bradley-Terry model based on a paired human preference dataset.
Alternatively, \citet{lin2025evaluating} uses a multi-modal large language models to exploit the knowledge of LLMs by performing visual question answering task to measure the alignment between texts and images. While existing reward models can operate as a proxy for ground-truth reward models, the inherent noise within the data due to finite capacity and coverage, inevitably affect the performance negatively when used for fine-tuning. Our work addresses the above-mentioned issues by introducing a calibration method that approximates the win-rate, rather than using the rewards. 

\noindent
{\bf Fine-tuning diffusion models with rewards.}
Numerous methods have been developed for fine-tuning T2I diffusion models with reward models~\citep{lee2023aligning, black2023training, fan2023dpok, deng2024prdp, clark2023directly, lee2025parrot, wallace2023diffusion}. By formulating discrete diffusion sampling process as a reinforcement learning problem, \citet{black2023training} and \citet{fan2023dpok} develop fine-tuning diffusion models with policy gradient algorithms. 
However, those methods are computationally expensive and the training processes are usually unstable. While \citet{deng2024prdp} presents a scheme to scale RL fine-tuning for large-scale prompt dataset, it is not clear whether this approach can be applied to large-scale diffusion models. Instead of using RL, \citep{clark2023directly} proposes to directly fine-tune diffusion models by using the gradients from reward models. Yet, those approaches can only be applied to differentiable reward models, and extending to large-scale reward models (\emph{e.g.}, LLMs) is computationally prohibitive in practice. Inspired by the success of DPO~\citep{rafailov2024direct}, \citet{wallace2023diffusion} introduce Diffusion-DPO, which can effectively alleviate the computational loads and can be applied to large-scale diffusion models~\citep{esser2024scaling}. Consequent works~\citep{yuan2024self, karthik2024scalable, li2024aligning, liang2024step, lee2024direct} built upon Diffusion DPO to enhance preference optimization or customization of large-scale T2I diffusion models.

\section{Preliminaries}
\label{sec:prelim}
We first describe the preliminaries on preference optimization for diffusion models before presenting our method. More details can be found in Appendix~\ref{appendix:add_desc}.

\vspace{1mm}
\noindent
{\bf Diffusion models.}
The denoising diffusion model~\citep{sohl2015deep, song2019generative, ho2020denoising, song2020score} consists of forward processes, which gradually add noise to the data, and reverse processes, which generate data from noise. The forward process of a data $\mathbf{x}$ at time $t\in[0,1]$ forms a distribution $q(\mathbf{x}_t|\mathbf{x})$, given by $\mathbf{x}_t=\alpha_t\mathbf{x}+\sigma_t\eps$, where $\eps\sim\noise$, and $\alpha_t, \sigma_t$ are noise schedules. 
Let $\lambda_t = \log (\alpha_t^2 / \sigma_t^2)$ be log signal-to-noise ratio (SNR), then we express the diffusion training objective as a weighted $\eps$-prediction loss as in~\citep{kingma2023understanding}:
\begin{equation}
\label{eq:diffloss}
    \mathcal{L}_{\textrm{DM}}(\mathbf{x}) = \mathbb{E}_{t\sim\mathcal{U}(0,1),\eps}\big[-w_t \lambda_t' \|\eps_\theta(\mathbf{x}_t;t)-\eps\|_2^2\big]\text{,} 
\end{equation}
where $w_t$ is a weighting function, and $\lambda_t' = \mathrm{d}\lambda / \mathrm{d}t$. Note that most of diffusion~\citep{karras2022elucidating} and flow matching~\citep{lipman2022flow} training objectives can be expressed in Eq.~\eqref{eq:diffloss} by choosing $w_t$ and $\lambda_t$.
The reverse process generates data by solving time-discretized SDE~\citep{song2020score} or ODE~\citep{song2020denoising, karras2022elucidating}, which gradually denoises Gaussian noise into data by using trained diffusion model. Text-to-image diffusion models~\citep{nichol2021glide, ramesh2021zero, ramesh2022hierarchical, saharia2022photorealistic, rombach2022high} are conditional diffusion models that use text embeddings $\mathbf{c}$ from text encoders~\citep{radford2021learning, raffel2020exploring} as condition to generate image from text input. 
In this work, we denote $\eps_\theta(\mathbf{x}_t;\mathbf{c},t)$ as T2I diffusion model, and $p_\theta(\cdot|\mathbf{c})$ as distribution of the generated data given prompt $\mathbf{c}$. 

\vspace{1mm}
\noindent
{\bf Reward models.}
Given an image $\mathbf{x}$ and a condition $\mathbf{c}$, a reward model $R(\mathbf{x}, \mathbf{c})$ is a function that measures an utility of the input. The common approach is Bradley-Terry (BT) model~\citep{bradley1952rank, ouyang2022training}, which defines the preference distribution for a triplet $(\mathbf{c}, \mathbf{x}, \mathbf{x}')$:
\begin{equation}\label{eq:bt}
    \mathbb{P}(\mathbf{x} \succ \mathbf{x}' | \mathbf{c}) \coloneqq \sigma\big( R(\mathbf{x}, \mathbf{c}) - R(\mathbf{x}', \mathbf{c})\big )\text{,}
\end{equation}
where $\sigma(u) = (1+\exp(-u))^{-1}$ is a sigmoid function.

\vspace{1mm}
\noindent
{\bf Diffusion preference optimization.} The goal of reward fine-tuning is to optimize the model $p_\theta$ that maximizes the expected reward of generated output, which comes with KL regularization to prevent over optimization:
\begin{equation}\label{eq:obj}
    \max_\theta~\mathbb{E}_{\mathbf{c}, \mathbf{x}\sim p_\theta(\mathbf{x}|\mathbf{c})}[R(\mathbf{x}, \mathbf{c})] - \beta D_{\textrm{KL}}\big(p_\theta(\cdot | \mathbf{c}) \| p_{\textrm{ref}}(\cdot|\mathbf{c})\big)\text{,}
\end{equation}
where $\beta$ is a hyperparameter that controls the divergence.
To solve Eq.~\eqref{eq:obj}, direct alignment methods, \emph{e.g.}, DPO~\citep{rafailov2024direct}, have been applied to diffusion models~\citep{wallace2023diffusion}. At its core, it uses the closed-form solution of Eq.~\eqref{eq:obj}, which is given by $p^*(\mathbf{x}|\mathbf{c}) \propto p_{\textrm{ref}}(\mathbf{x}|\mathbf{c})\exp\big(\tfrac{1}{\beta}R(\mathbf{x}, \mathbf{c})\big)$. By replacing $p^*$ with $p_\theta$ and rearranging for $r$, applying Eq.~\eqref{eq:bt} for a ranked data pair $(\mathbf{x}^+, \mathbf{x}^-)$ gives us following general preference optimization objective~\citep{tang2024generalized}:
\begin{equation}\label{eq:dpoloss}
    \ell(\theta) = g \bigg(\beta\log\frac{p_\theta(\mathbf{x}^+|\mathbf{c})}{p_{\textrm{ref}}(\mathbf{x}^+|\mathbf{c})} - \beta\log\frac{p_\theta(\mathbf{x}^-|\mathbf{c})}{p_{\textrm{ref}}(\mathbf{x}^-|\mathbf{c})}\bigg)\text{,}
\end{equation}
where $g$ is any convex loss function, \emph{e.g.}, $g(u) =-\log \sigma(u)$ gives us DPO~\citep{rafailov2024direct} objective, and $g(u) = (1-u)^2$ gives us identity preference optimization (IPO)~\citep{azar2024general} objective.

However, directly applying Eq.~\eqref{eq:dpoloss} to diffusion models is not straightforward as the log-likelihoods of diffusion models are intractable. \citet{wallace2023diffusion} propose a method to compute log-ratio and derive DPO loss for diffusion models by marginalizing the log-ratio through forward process $q(\mathbf{x}_{0:1}|\mathbf{x})$ to compute the log-ratio with $\eps$-prediction losses: 
\begin{equation*}\label{eq:diffratio}
    \mathbb{E}_{q(\mathbf{x}_{0:1}|\mathbf{x})}\bigg[\log \frac{p_\theta(\mathbf{x}_{0:1}|\mathbf{c})}{p_{\textrm{ref}}(\mathbf{x}_{0:1}|\mathbf{c})}\bigg] = \mathbb{E}_{t,\eps}\big[ R_\theta(\mathbf{x}_t, \mathbf{c}, t)\big]\text{,}
\end{equation*}
where  $R_\theta(\mathbf{x}_t, \mathbf{c}, t)  \! = \!  \lambda_t'\big(\|\eps_\theta(\mathbf{x}_t;\mathbf{c},t) -$$ \eps\|_2^2 - \|\eps_{\textrm{ref}}(\mathbf{x}_t;\mathbf{c},t) -\eps\|_2^2\big)$ 
for $\mathbf{x}_t=\alpha_t\mathbf{x}+\sigma_t\eps$ with $t\sim\mathcal{U}(0,1)$ and $\eps\sim\noise$. 
By applying this to Eq.~\eqref{eq:dpoloss} and taking the expectation out of $g$ yields diffusion preference optimization objective:
\begin{equation}\label{eq:dffpo}
    \mathcal{L}(\theta) = \mathbb{E}_{t,\eps^+, \eps^-}\big[g\big(\beta R_\theta(\mathbf{x}_t^+, \mathbf{c}, t) - \beta R_\theta(\mathbf{x}_t^-,\mathbf{c}, t) \big)\big]\text{,}
\end{equation}
where $\mathbf{x}_t^+ \! =  \alpha_t\mathbf{x}^+ + \sigma_t\eps$, $\mathbf{x}_t^-=\alpha_t\mathbf{x}^- + \sigma_t\eps^-$ for $t\sim\mathcal{U}(0,1)$ and $(\eps^+, \eps^-)\sim \noise \times \noise $.

\section{Proposed Method}
In this section, we introduce our method for calibrated preference optimization. We refer to Fig.~\ref{fig:overview} for the overview.

\subsection{Motivation}\label{sec:problem}
The challenges in multi-reward optimization is in achieving the Pareto optimality among reward signals, especially even when they conflict. For example, when optimizing models for image aesthetics, it often results in reduced image-text alignment as aesthetic reward models do not consider textual information (\emph{e.g.}, see Tab.~\ref{tab:single}). One common practice is to use weighted sum of rewards as a proxy for the total reward function~\citep{clark2023directly}. However, those rigid formulations cannot effectively consider all aspects of utilities, which might lead to suboptimal performance, \emph{e.g.}, biased towards certain reward signals. Another approach is using the rewarded soups~\citep{rame2024rewarded}, which merges the independently reward fine-tuned model with model soup~\citep{wortsman2022model}. Nevertheless, optimizing for a single reward is prone to reward over-optimization~\citep{gao2023scaling, rafailov2024scaling} and result in significant performance loss.

Our core assumption is that the difficulties in multi-reward optimization lie in the inconsistency between the black-box distribution of rewards. To address this challenge, we propose calibrated preference optimization to minimize inconsistencies by fine-tuning with general and unified metrics. In the following, we provide details of our method.

\subsection{CaPO}\label{sec:capo}
Although we consider reward models as a proxy to represent the utility of a sample, directly using the reward values can lead to unsatisfactory results if they are not properly calibrated. 
Specifically, when using Bradley-Terry model~\citep{bradley1952rank}, the reward value often does not measure the goodness of a sample, even though the model exhibits high prediction accuracy in classifying the human preference. 
Furthermore, the varying range of reward becomes problematic when using multiple reward signals, making it difficult to obtain balanced updates. 

\vspace{1mm}
\noindent
{\bf Calibrated rewards.}
To address these issues, we propose to use expected win-rate as a unified measure for maximization target. Formally, let $\mathbb{P}(\mathbf{x} \succ \mathbf{x}' | \mathbf{c})$ be a win-rate of data $\mathbf{x}$ over $\mathbf{x}'$ with prompt $\mathbf{c}$. We define the win-rate of data $\mathbf{x}$ over a distribution $p(\cdot | \mathbf{c})$:
\begin{equation}\label{eq:winrate}
    \mathbb{P}(\mathbf{x} \succ p | \mathbf{c}) \coloneqq \mathbb{E}_{\mathbf{x}' \sim p(\cdot | \mathbf{c})}\big[\mathbb{P}(\mathbf{x} \succ \mathbf{x}'|\mathbf{c})\big]\text{.}
\end{equation}
As our goal is to improve over reference model $p_{\textrm{ref}}$, we consider $\mathbb{P}(\mathbf{x} \succ p_{\textrm{ref}}|\mathbf{c})$ as our target of interest. 
By using expected win-rate over reference model, we directly seek for improvement over a pretrained model, which quantifies the general goodness of a data. Furthermore, the bounded range makes it more favorable for multi-reward optimization. Since the expected win-rate is not available in general, we approximate it through averaging the pairwise win-rate computed by a reward model. Suppose we generate $N$ batch of samples $\{\mathbf{x}_i\}_{i=1}^N$ from $p_{\textrm{ref}}(\cdot|\mathbf{c})$, then we define \emph{calibrated reward} $R_{\textrm{ca}}(\mathbf{x}_i, \mathbf{c})$ for each sample $i$:
\begin{equation}\label{eq:calre}
\vspace{-3mm}
    R_{\textrm{ca}}(\mathbf{x}_i, \mathbf{c}) = \frac{1}{N-1} \sum_{j\neq i} \sigma \big( R(\mathbf{x}_i, \mathbf{c}) - R(\mathbf{x}_j, \mathbf{c}) \big)\text{,}
\end{equation}
where we have $R_{\textrm{ca}}(\mathbf{x}, \mathbf{c}) \approx \mathbb{P}(\mathbf{x}\succ p_{\textrm{ref}}|\mathbf{c})$ for large $N$.

\vspace{1mm}
\noindent
{\bf CaPO loss.}
We replace $R(\mathbf{x}, \mathbf{c})$ in Eq.~\eqref{eq:obj} with $R_{\textrm{ca}}(\mathbf{x}, \mathbf{c})$, and introduce calibrated preference optimization objective that fine-tunes the model to maximize the calibrated reward. Similar to Eq.~\eqref{eq:dffpo}, we define CaPO loss by matching the difference of the calibrated rewards with regression loss~\citep{deng2024prdp, fisch2024robust}, which also guarantees the optimality condition. 
Specifically, given data pair $(\mathbf{x}^+, \mathbf{x}^-)$, we define CaPO objective:
\vspace{-1mm}
\begin{align}\label{eq:capoloss}
\begin{split}
    \mathcal{L}_{\textrm{CaPO}}(\theta) &= \underset{t,\eps, \eps'}{\mathbb{E}}\bigg[
    \bigg( R_{\textrm{ca}}(\mathbf{x}^+, \mathbf{c}) - R_{\textrm{ca}}(\mathbf{x}^-, \mathbf{c}) \\
    &- \beta \big(R_\theta(\mathbf{x}_t^+,\mathbf{c},t) - R_\theta(\mathbf{x}_t^-, \mathbf{c}, t)\big)\bigg)^2\bigg]\text{,}
\end{split}
\end{align}
where $\mathbf{x}_t^+=\alpha_t\mathbf{x}^+ + \sigma_t\eps^+$, $\mathbf{x}_t^-=\alpha_t\mathbf{x}^- + \sigma_t\eps^-$, for $t\sim\mathcal{U}(0,1)$ and $(\eps^+, \eps^-)\sim \noise \times \noise$. Note that CaPO is a special case of Eq.~\eqref{eq:dffpo} with $g(u)=(\Delta R - u)^2$, where $\Delta R = R^+ - R^- $ is a difference between calibrated rewards. Thus, CaPO is a generalization of IPO~\citep{azar2024general}, which strictly assign $\Delta R = 1$ for all pairs. Compared to IPO, CaPO assigns a dynamic target for the preference learning, which helps maximizing the gain without reward over-optimization. 

\subsection{Preference Pair Selection}
\label{sec:frs}
The best-of-$N$ sampling~\citep{nakano2021webgpt, cobbe2021training} or rejection sampling~\citep{touvron2023llama} methods that 
select samples with highest reward from $N$ generation
are commonly used in RLHF. For a single reward, it is straightforward to choose the sample $\mathbf{x}^+$ with highest reward, and $\mathbf{x}^-$ that has lowest reward to maximize the margin between the pair. For multi-reward optimization, the na\"ive approach is to use weighted sum as the total proxy reward model, and perform rejection sampling with it. However, choosing the weights often relies on heuristics, and the optimal weights might be dynamic depending on the input, which can lead to suboptimal performance.

In order to achieve the Pareto optimal solution, we propose \emph{frontier-based rejection sampling} (FRS), which selects the set of positive samples $X^+(\mathbf{c})$ and negative samples $X^-(\mathbf{c})$ for each prompt $\mathbf{c}$ by finding Pareto optimal set. Specifically, we use a non-dominated sorting algorithm~\citep{deb2002fast} to find the upper and lower Pareto frontier. The goal of FRS is to push apart from the lower Pareto frontier and pull towards the upper Pareto frontier, which helps to achieve Pareto optimality. Given $L$ reward models, let $R_{\textrm{ca}}^{(j)}$ be $j$-th calibrated rewards for $j=1,\ldots, L$, then we define $\mathbf{x}$ dominates $\mathbf{x}'$ if and only if $R_{\textrm{ca}}^{(j)}(\mathbf{x}, \mathbf{c}) \geq R_{\textrm{ca}}^{(j)}(\mathbf{x}', \mathbf{c})$ for all $j=1,\ldots, L$. Then finding a set of non-dominated data points is referred as finding Pareto set, which forms an upper frontier. Conversely, one can define a set of dominated data that forms a lower frontier. 
After removing the potential duplicates of non-dominated and dominated sets, we take $X^+(\mathbf{c})$ by filtered non-dominated sets and $X^-(\mathbf{c})$ by set of dominated set. Given positive set ${X}^+(\mathbf{c})$ and $X^-(\mathbf{c})$, we sample a positive sample $\mathbf{x}^+\sim X^+(\mathbf{c})$ and $\mathbf{x}^-\sim X^-(\mathbf{c})$ to construct a pair. We use CaPO loss to update the model with ensemble of calibrated rewards for optimization target:
\vspace{-2mm}
\begin{equation*}
\vspace{-2mm}
    R_{\textrm{ca}}(\mathbf{x}, \mathbf{c}) = \frac{1}{L} \sum_{j=1}^L R_{\textrm{ca}}^{(j)}(\mathbf{x}, \mathbf{c})\text{,}
\end{equation*}
and use CaPO loss in Eq.~\eqref{eq:capoloss} for the update.

\subsection{Loss weighting}\label{sec:adv}
The choice of log-SNR $\lambda_t$ and weighting function $w_t$ has large impact on the generation quality and convergence of diffusion model pretraining. Intuitively, when $\lambda_t$ is large, \emph{i.e.}, small amount of noise is added, the denoising task becomes easier, and conversely the task becomes harder as $\lambda_t$ becomes smaller, thus weighting function as a monotonically decreasing weighting function of $\lambda_t$ seems a reasonable choice. In \citep{kingma2023understanding}, those monotonic weighting are theoretically shown to be the weighted evidence lower bound (ELBO), and demonstrated better quality than the non-monotonic counterpart. In this work, we also propose to use monotonic loss weighting to our CaPO loss, which is equivalent to regularizing with weighted ELBO instead of KL divergence in Eq.~\eqref{eq:obj}. Specifically, we apply sigmoid weighting with bias, \emph{i.e.}, $w_t = w(\lambda_t) = \sigma(-\lambda_t + b)$, where $b$ is a bias hyperparameter~\citep{kingma2023understanding, hoogeboom2024simpler}. See supplementary for details.

\begin{table}[t]
\centering
\small
\vspace{-0.1in}
\centering\small
\newcolumntype{a}{>{\columncolor{mygreen! 10}}c}

\setlength\tabcolsep{1.7pt} 
\begin{tabular}{l | acc | cac | cca }
\toprule
 & MPS & VQA & VILA  & MPS & VQA & VILA & MPS & VQA & VILA  \\
\midrule
DPO  & 58.5  &49.3 & 61.7 & 53.1 & 50.6  & 55.9  & 52.6 & 46.4 & 81.8\\
IPO & 56.8 & {\bf 50.1} & 64.1 & 53.1 & 51.9  & 53.8  & 53.3 & 48.5 & 76.1\\
CaPO & {\bf 61.1} & 49.7 & {\bf 64.9} & {\bf 55.5} & {\bf 53.2}  & {\bf 58.7}  & {\bf 54.1} & {\bf 49.6} & {\bf 83.1}\\
\bottomrule
\end{tabular}
\vspace{-3mm}
\caption*{(a) Base model SDXL}
\begin{tabular}{l | acc | cac | cca }
\toprule
 & MPS & VQA & VILA  & MPS & VQA & VILA & MPS & VQA & VILA  \\
\midrule
DPO  & 55.2  & 53.2 & 54.4 
     & 52.1  & 53.2 & 52.9 
     & 53.1 & 48.7  & 70.1 \\
IPO  & 51.1  & 52.1 & 48.3 
     & 52.8  & 51.9 & 51.1  
     & {\bf 58.3} & 50.2 & 70.8  \\
CaPO & {\bf 58.1} & {\bf 53.3} & {\bf 63.4} 
     & {\bf 54.4} & {\bf 55.4} & {\bf 59.4} 
     & {57.4} & {\bf 50.8} & {\bf 74.0} \\
\bottomrule
\end{tabular}
\vspace{-3mm}
\caption*{(b) Base model SD3-M}
\vspace{-2mm}
\caption{
\textbf{Single reward results.} We report the win-rate (\%) over base model by using automatic evaluation with each reward model. We use Parti prompts~\citep{yu2022scaling} and DPG-bench prompts~\citep{hu2024ella} to generate images for each SDXL, and SD3-M models, respectively. We highlight the \textcolor{mygreen}{column} to indicate the rewards used for fine-tuning.
}\label{tab:single}
\vspace{-5mm}
\end{table}

\begin{table*}[t]
\centering
\small
\centering\small
\newcolumntype{a}{>{\columncolor{mygreen! 10}}c}
\setlength\tabcolsep{1.5pt} 
\begin{minipage}{0.49\textwidth}
\centering\small
\begin{tabular}{ll  cc cc cc  }
\toprule

&& \multicolumn{2}{c}{MPS}  & \multicolumn{2}{c}{VQAscore}  & \multicolumn{2}{c}{VILA} \\
\cmidrule(lr){3-4}\cmidrule(lr){5-6} \cmidrule(lr){7-8}
Objective & Method  & Win (\%) & Score & Win (\%) & Score & Win (\%) & Score\\

\midrule
SDXL & \multicolumn{1}{c}{-} & - & 11.30 & - & 0.826 & - &5.953 \\
\midrule
\multirow{3}{*}{DPO} 
& {\tt{SUM}}  & 57.2 & 11.48 & 52.1 & 0.829 & 71.9 & 6.193  \\
& {\tt{SOUP}} & 56.5 & 11.46 & 52.2 & 0.830 & 74.3 & 6.227 \\
& {\tt{FRS}}  & 58.1 & 11.54 & 52.9 & 0.834 & 78.6 & 6.294 \\
\midrule
\multirow{3}{*}{IPO} 
& {\tt{SUM}}  & 57.4 & 11.49 & 51.1 & 0.828 & 66.8 & 6.111  \\
& {\tt{SOUP}} & 55.4 & 11.44 & 52.0 & 0.830 & 70.3 & 6.154  \\
& {\tt{FRS}}  & 57.8 & 11.52 & 52.0 & 0.830 & 74.4 & 6.238  \\
\midrule
\multirow{3}{*}{CaPO} 
& {\tt{SUM}}  & {\bf 61.2} & 11.62 & 52.5 & 0.834 & 75.0 & 6.258 \\
& {\tt{SOUP}} & 59.4 & 11.44 & 52.8 & 0.835 & 77.6 & 6.259 \\
\rowcolor{mygreen!10}
& {\tt{FRS}}  & {\bf 61.2} & {\bf 11.66} & {\bf 54.6} & {\bf 0.839} & {\bf 79.2} & {\bf 6.340} \\
\bottomrule
\end{tabular}
\caption*{(a) Base model SDXL}
\end{minipage}
\hfill
\begin{minipage}{0.49\textwidth}
\centering\small
\hfill
\begin{tabular}{ll  cc cc cc  }
\toprule
&& \multicolumn{2}{c}{MPS}  & \multicolumn{2}{c}{VQAscore}  & \multicolumn{2}{c}{VILA} \\
\cmidrule(lr){3-4}\cmidrule(lr){5-6} \cmidrule(lr){7-8}
Objective & Method  & Win (\%) & Score & Win (\%) & Score & Win (\%) & Score\\
\midrule
SD3-M & \multicolumn{1}{c}{-} & - & 13.39 & - & 0.908 & - & 5.793\\
\midrule
\multirow{3}{*}{DPO} 
& {\tt{SUM}} & 55.3 & 13.50 & 52.8 & 0.910 & 55.0 & 5.832 \\
& {\tt{SOUP}} & 56.1 & 13.39 & 54.7 & 0.908 & 63.4 & 5.875 \\
& {\tt{FRS}} & 56.7 & 13.55 & 53.2 & 0.909 & 68.7 & 5.922 \\
\midrule
\multirow{3}{*}{IPO} 
& {\tt{SUM}}  & 54.1 & 13.47 & 53.9 & 0.912 & 58.9 & 5.847 \\
& {\tt{SOUP}} & 55.6 & 13.39 & 53.5 & 0.910 & 60.4 & 5.848 \\
& {\tt{FRS}}  & 55.5 & 13.55 & 54.6 & 0.913 & 64.7 & 5.913 \\
\midrule
\multirow{3}{*}{CaPO} 
& {\tt{SUM}}  & 57.8 & 13.56 & 54.3 & 0.912 & 57.0 & 5.833 \\
& {\tt{SOUP}} & {\bf 59.4} & {\bf 13.60} & 54.9 & 0.911 & 67.6 & 5.896 \\
\rowcolor{mygreen!10}
& {\tt{FRS}}  & 59.0 & 13.58 & {\bf 55.7} & {\bf 0.914} & {\bf 69.3} & {\bf 5.943} \\
\bottomrule
\end{tabular}
\caption*{(b) Base model SD3-M}
\end{minipage}
\vspace{-2mm}
\caption{
\textbf{Multi-reward results.} We report the average reward scores (Score) and win-rate (\%) over base model by using automatic evaluation with each reward model (Win).
We compare preference objectives DPO~\cite{wallace2023diffusion}, IPO~\cite{azar2024general}, and CaPO and combination with different pair selection methods, \emph{e.g.}, using sum of rewards to conduct top-1 and worst-1 sampling ({\tt{SUM}}), and using frontier-based rejection sampling ({\tt{FRS}}).
Furthermore, we compare our method with rewarded soup~\citep{rame2024rewarded}, by merging single reward optimized models ({\tt{SOUP}}). 
}\label{tab:multi}
\end{table*}

\section{Experiments}
\label{sec:experiment}
\begin{figure*}[t]
    \small\centering
    \includegraphics[width=1.0\textwidth]{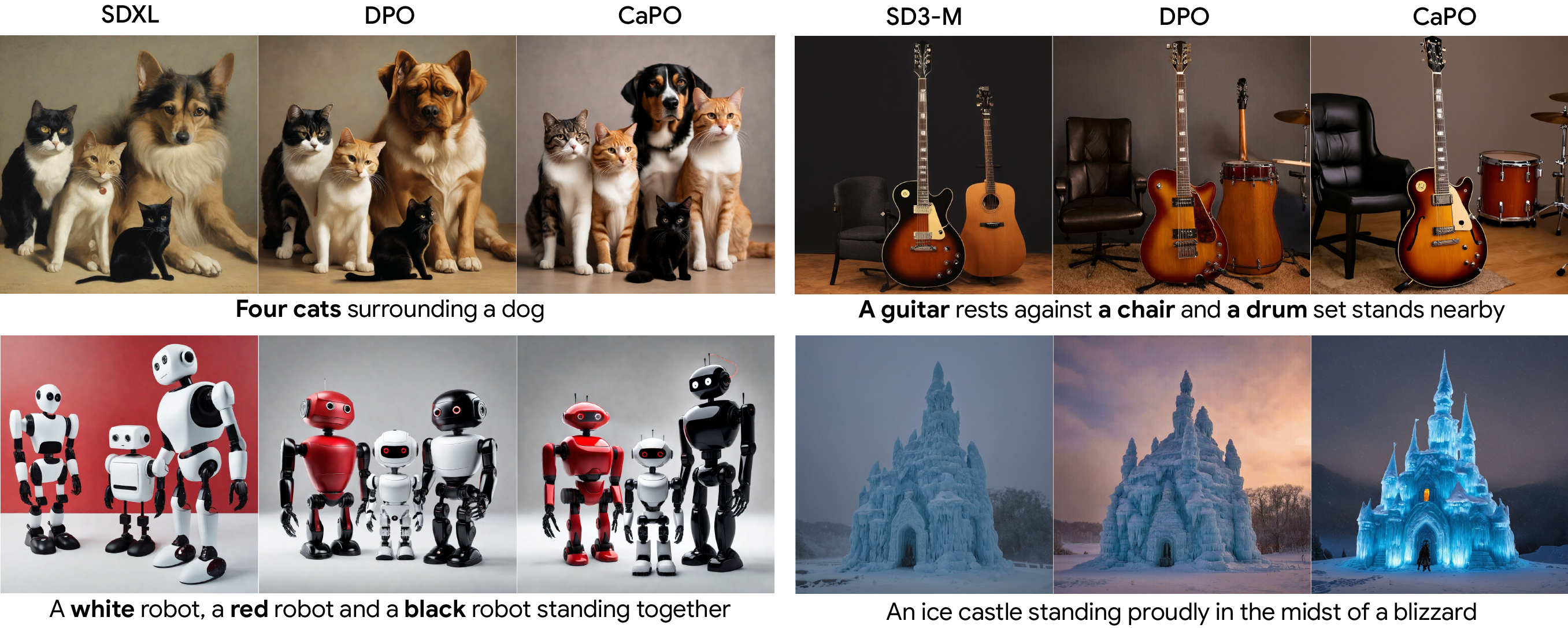}
    \caption{
    \textbf{Qualitative comparison}. We present qualitative comparison of using multiple rewards using our frontier rejection sampling method when fine-tuning with CaPO objective.}
    \label{fig:qual_comparison}
\end{figure*}
\begin{table*}[t]
\centering
\small
\vspace{-0.1in}
\centering\small
\setlength\tabcolsep{2pt}
\begin{tabular}{l ccccccc c cccccc}
\toprule
      & \multicolumn{7}{c}{GenEval} && \multicolumn{6}{c}{T2I-Compbench}\\
      \cmidrule(lr){2-8}
      \cmidrule(lr){10-15}
Model & Single & Two & Counting & Colors & Position & \makecell{Color \\ Attribution} & Overall  && Color & Shape & Texture & Complex & Spatial & Non Spatial\\
\midrule
FLUX-dev   & 0.98 & 0.84 & {0.67} & 0.76 & 0.23 & 0.46 & 0.66
&& 0.740 & 0.486 & 0.650 & 0.477 & 0.220 & 0.306 \\
FLUX-schnell & {0.99} & 0.81 & 0.58 & 0.63 & 0.25 & 0.35 & 0.60 
&& 0.642 & 0.509 & 0.646 & 0.424 & 0.185 & 0.304  \\
SD3.5-L & {0.99} & {0.89} & 0.62 & 0.82 & 0.26 & 0.53 & 0.69
&& 0.763 & {0.602} & {0.766} & {0.520} & 0.219 & {0.314} \\
\midrule
SDXL     & 0.98 & 0.74 & 0.39 & 0.85 & 0.15 & 0.23 & 0.55 
&& 0.592 & 0.500 & 0.608 & 0.465 & 0.159 & 0.312 \\
\rowcolor{mygreen!10}
CaPO+SDXL & {0.99} & 0.79 & 0.48 & {0.86} & 0.15 & 0.28 & 0.59 
&& 0.646 & 0.537 & 0.633 & 0.491 &0.172 & 0.312 \\
\midrule
SD3-M & {0.99} & 0.84 & 0.56 & 0.84 & {0.32} & 0.52 & 0.68 
&& 0.775 & 0.546 & 0.712 & 0.505 & 0.221 & 0.309 \\
\rowcolor{mygreen!10}
CaPO+SD3-M & {0.99} & { 0.87} & 0.63 & {0.86} & 0.31 & {0.59} & {0.71} 
&& {0.788} & {0.572} & {0.731} & {0.509} & {0.230} & {0.313} \\
\bottomrule
\end{tabular}
\caption{
\textbf{T2I Benchmarks evaluation.}
We compare the benchmark results of CaPO-SDXL and CaPO-SD3-M on text-to-image benchmarks, \emph{e.g.}, GenEval~\citep{ghosh2024geneval} and T2I-Compbench~\citep{huang2023t2i}, with various open-source state-of-the-art models (\emph{e.g.}, Flux-dev~\citep{flux2024}, and Flux-schnell~\citep{flux2024}, and SE3.5-L~\citep{esser2024scaling}).
We observe that with CaPO, the majority of evaluation metrics for SDXL and SD3-M show improvement. For comparison, we also include the most recent three image generation models, which are 3$\times$ larger compared to SDXL and SD3-M. 
}\label{tab:geneval}
\vspace{-0.05in}
\end{table*}

\vspace{0.05in}
\noindent
{\bf Models and datasets.}
We use Stable Diffusion XL (SDXL) \citep{podell2023sdxl} and Stable Diffusion 3 medium (SD3-M) \citep{esser2024scaling} as our base text-to-image diffusion models in all experiments. To collect the training dataset, we use 100K prompts from DiffusionDB~\citep{wang2022diffusiondb}, and generate $N=16$ images per prompt. We also experiment with using 8, or 32 images per prompt, and select 16 images per prompt, which provides a good trade-off between computational cost and performance improvement.
To generate images, we use DDIM \citep{song2020denoising} sampler with guidance scale 7.5 for 50 steps, and flow DPM-solver~\citep{lu2022dpm} with guidance scale 5.0 for 50 steps for each SDXL and SD3-M, respectively.
Note that we only use the images generated by the same diffusion model for experiments. We refer to supplementary for detailed experimental setup.

\vspace{1mm}
\noindent
{\bf Reward models.}
We consider three reward models that cover diverse aspects of the T2I generation. For general human preference (\emph{i.e.}, overall quality), we use MPS score~\citep{zhang2024learning}, which is a state-of-the-art reward model for human preference. For image-text alignment, we use VQAscore~\citep{lin2025evaluating}, which uses a vision-language model (CLIP-FlanT5-XXL) to compute scores by performing visual question answering tasks. Specifically, VQAscore measures the probability $P(\texttt{Yes}|\mathbf{x}, \texttt{Q}(\mathbf{c}))$ by using the output logits of the model, where $\texttt{Q}$ is a template for the question. We also use VILA score~\citep{ke2023vila} pretrained on AVA~\cite{murray2012ava} dataset to evaluate image aesthetics. While VQAscore and VILA score are not trained with BT model, we adjust to approximate with BT model (see supplementary for detail).

\subsection{Single reward experiments}
\label{sec:single_exp}
\vspace{1mm}
\noindent
{\bf Experimental setups.} We evaluate CaPO against state-of-the-art preference learning objectives such as DPO~\citep{wallace2023diffusion} and IPO~\citep{azar2024general} for diffusion models. For each method, we train with three reward models (MPS, VQAscore, and VILA) by selecting the top--1 and worst--1 pair.
For evaluation, we use Parti prompts~\citep{yu2022scaling} to generate images for SDXL fine-tuned model and DPG-Bench~\citep{hu2024ella} prompts to generate images for SD3-M fine-tuned model. We report the win-rate against the base model using each reward model.

\vspace{0.05in}
\noindent
{\bf Results.} Tab.~\ref{tab:single} shows that CaPO achieves the highest win-rate for each reward model used for fine-tuning, as well as other reward models. Especially, when using the VILA model for training, DPO shows significant drop in VQAscore, while showing comparable performance with CaPO in VILA score.
On the other hand, IPO shows better robustness than DPO in reward hacking, but the gain of the performance is lower than DPO and CaPO in general. We notice that even though we optimized for a reward, other rewards also increase at some cases. This is partially due to the inherent correlation residing in reward models, \emph{e.g.}, increasing MPS score results in increase in VILA score, as image aesthetics is an important factor in overall quality. 

\subsection{Multi-reward experiments}
\label{sec:exp_main}
\vspace{0.05in}
\noindent
{\bf Experimental setups.}
We consider MPS, VQAscore, and VILA scores for multi-reward experiments. For baselines, we evaluate CaPO against DPO and IPO as in Sec.~\ref{sec:single_exp}. Furthermore, we conduct experiments on different methods in adapting for multiple rewards. Specifically, we compare frontier-based rejection sampling (FRS) (\emph{i.e.}, Sec.~\ref{sec:frs}) with sum-of-rewards (Sum), and merging the models fine-tuned with single reward (\emph{i.e.}, model soup~\citep{wortsman2022model}). For sum-of-rewards, we directly add the calibrated rewards, and perform top-1 and worst-1 pair selection for training data. For model soup, we re-use fine-tuned models from Sec.~\ref{sec:single_exp}, and use spherical linear interpolation~\citep{shoemake1985animating, rame2024warp} to merge models, which performs slightly better than linear interpolation in our experiments. We use uniform weights (\emph{i.e.}, 1/3 each) for both sum-of-rewards and model soup. For evaluation, we generate images using Parti prompt dataset~\citep{yu2022scaling} and DPG-bench prompt dataset for SDXL and SD3-M, respectively. For evaluation, we report the average reward scores, and the win-rate against the base model by using each reward model.

\vspace{1mm}
\noindent
{\bf Quantitative results.}
Tab.~\ref{tab:multi} shows the results. First, joint training of multiple rewards by using frontier-based rejection sampling consistently outperforms pair selection with sum of rewards on all preference optimization objectives. While model merging (Soup) shows comparable performance to FRS when using DPO and IPO objectives for training, using CaPO objective with FRS outperforms model soup of CaPO fine-tuned models with single reward. When comparing CaPO, DPO, and IPO, CaPO with FRS shows higher win-rates and average rewards compared to DPO and IPO with FRS, which is consistent with the results of Tab.~\ref{tab:single}.

\vspace{1mm}
\noindent
{\bf Qualitative results.}
In Fig.~\ref{fig:qual_comparison}, we provide qualitative comparison of our method on SDXL and SD3-M, compared to DPO trained with multi-reward frontier-based rejection sampling. While both DPO and CaPO shows improved image aesthetics such as contrast or color compared to base SDXL, we see that CaPO demonstrates better image-text alignment and aesthetic quality compared to DPO, following the quantitative results in Tab.~\ref{tab:multi}. We refer to supplementary for additional examples.

\vspace{1mm} 
\noindent
{\bf Benchmark results.}
For quantitative analysis, we evaluate our models on various T2I benchmarks; GenEval~\citep{ghosh2024geneval} which evaluates object-focused generation, and T2I-Compbench~\citep{huang2023t2i} for compositional generation. We compare our method with the base models SDXL and SD3-M, as well as open-source state-of-the-art T2I diffusion models such as FLUX-dev~\citep{flux2024}, FLUX-schnell~\citep{flux2024}, and Stable Diffusion 3.5-large (SD3.5-L)~\citep{esser2024scaling}. Tab.~\ref{tab:geneval} shows that CaPO improves the performance of the base model, \emph{e.g.}, 0.55$\rightarrow$0.59 for SDXL, 0.68$\rightarrow$0.71 on GenEval overall score, and on almost every metrics in T2I-Compbench. 

\begin{table}[t]
\centering
\small
\vspace{-0.1in}
\centering\small
\newcolumntype{a}{>{\columncolor{mygreen! 10}}c}

\begin{tabular}{l  ccccc}
\toprule
 & Pickscore & MPS & VQA & VILA\\
\midrule
Diffusion-DPO~\citep{wallace2023diffusion} & 22.71 & 11.59 & 0.834 & 6.049 \\
DPO-Syn  & 22.74 & 11.59 & 0.825 & 6.074 \\
CaPO  & {\bf 22.83} & {\bf 11.71} & {\bf 0.838} & {\bf 6.141}\\
\bottomrule
\end{tabular}
\vspace{-2mm}
\caption{
\textbf{Comparison with Diffusion-DPO~\citep{wallace2023diffusion}.} 
We compare CaPO with Diffusion-DPO, which is trained on human annotated preference dataset Pick-a-pic~\citep{kirstain2023pick}. 
For fair comparison, we train CaPO with same prompts from Pick-a-pic, but trained with generated images from SDXL. Also, we use Pickscore~\citep{kirstain2023pick}, which is trained on Pick-a-pic dataset. We also train DPO for our synthetic dataset, denoted as DPO-Syn.
We report Pickscore, MPS, VQAscore, and VILA score by generating images from Parti prompts. 
}\label{tab:dpo_comp}
\vspace{-0.05in}
\end{table}

\begin{figure}[t]
    \small\centering
    \includegraphics[width=0.99\columnwidth]{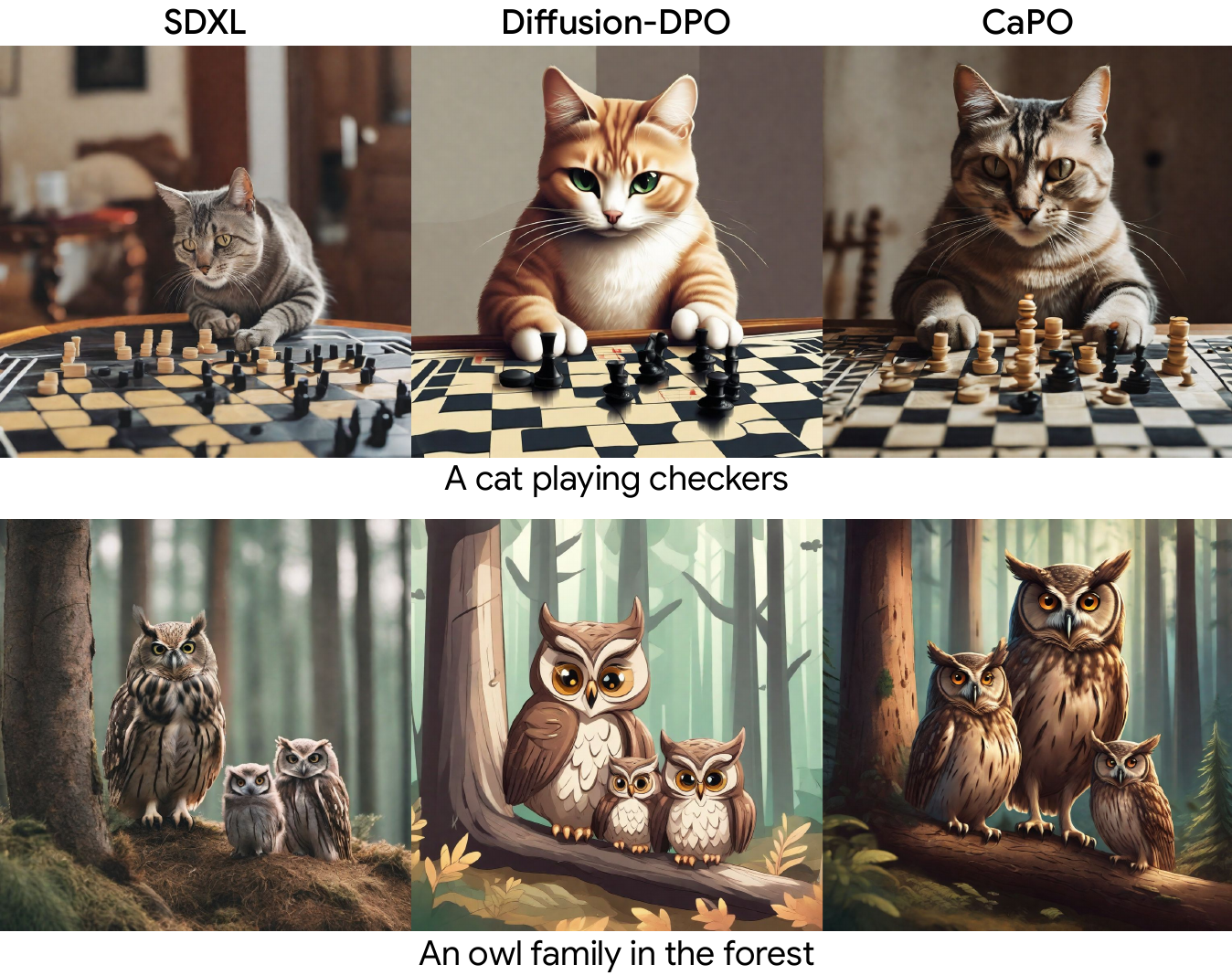}
    \caption{
    \textbf{Qualitative comparison with Diffusion-DPO~\citep{wallace2023diffusion}.} 
    We show qualitative comparison between SDXL, Diffusion-DPO, and CaPO using Pickscore~\citep{kirstain2023pick}. 
    }
    \label{fig:diffdpocomp}
    \vspace{-10pt}
\end{figure}

\subsection{Ablation Studies}
\noindent
{\bf Comparison with Diffusion-DPO~\citep{wallace2023diffusion}.} 
We compare our method with Diffusion-DPO~\citep{wallace2023diffusion}, which fine-tunes SDXL on the human preference dataset Pick-a-pic~\citep{kirstain2023pick}.
For fair comparisons, we use the same 58K prompts in Pick-a-pic v2 dataset, and use Pickscore~\citep{kirstain2023pick}, which is a reward model trained on Pick-a-pic dataset, as our reward signal.
Here, we generate $N=16$ images for each prompt, and select a pair by choosing highest and lowest reward, following Sec.~\ref{sec:single_exp}. We also train DPO on our synthetic data (DPO-Syn), to show the effect of synthetic data for fine-tuning. 
For evaluation, we generate images using Parti prompts~\citep{yu2022scaling}, and compare Pickscore, MPS, VQAscore, and VILA scores. Tab.~\ref{tab:dpo_comp} shows the results. 
Note that while DPO-Syn scores higher than Diffusion-DPO on Pickscore and VILA score, Diffusion-DPO outperforms on VQAscore. 
On the other hand, CaPO strictly shows better performance than Diffusion-DPO. In Fig~\ref{fig:diffdpocomp}, we show visual comparison between SDXL, Diffusion-DPO, and CaPO, which shows consistent trends as in Tab.~\ref{tab:dpo_comp}.

\vspace{1mm}
\noindent
{\bf Effect of loss weighting.} 
We demonstrate the effect of loss weighting that we proposed in Sec.~\ref{sec:adv}. Specifically, we compare the performance of CaPO when using sigmoid loss weighting, and without loss weights (\emph{i.e.}, $w_t\lambda_t' = -1$). We vary the bias of loss weight by $b=1.0, 1.5, 2.0$. Tab.~\ref{tab:abl} shows the results of SDXL CaPO models trained with multi-reward experimental setup. We notice using loss weighting significantly improves the performance, while the best $b$ achieves at $b=1.5$. Note that the trend of bias differs for SD3-M, which we refer to supplementary for details.

\begin{table}[t]
\centering
\small
\vspace{-0.1in}
\centering\small
\begin{tabular}{l  ccc}
\toprule
 & MPS & VQA & VILA \\
\midrule
Constant weighting  & 56.5  & 51.8 & 70.8  \\
\midrule
Sigmoid weighting ($b=1.0$)  & 59.1 &	54.5 &73.3 \\
Sigmoid weighting ($b=1.5$)  & 61.2 & 54.6 & 79.2   \\
Sigmoid weighting ($b=2.0$)  & 58.6 & 52.6 & 75.2   \\
\bottomrule
\end{tabular}
\vspace{-2mm}
\caption{
\textbf{Ablation on loss weighting.} We show the results of CaPO multi-reward fine-tuning SDXL with constant weighting (\emph{i.e.}, $-w_t\lambda_t'=1$), and sigmoid weighting by varying bias $b=1.0, 1.5, 2.0$. Using sigmoid weighting shows better results than constant weighting, and $b=1.5$ performs the best.
}\label{tab:abl}
\vspace{-5mm}
\end{table}

\subsection{Discussions}
While our method can improve the quality of T2I generation, some of the improvements (\emph{e.g.}, improving the text rendering for SDXL) is difficult, which is bounded by the performance of original model. However, for more powerful diffusion models (\emph{e.g.}, SD3), we show that our method can improve the text rendering as well. Furthermore, our approach is built upon offline data generation, which often suffers from slow convergence. Extending CaPO to online learning problems is a promising direction and we leave it for future work.
\section{Conclusion}
\label{sec:conclusion}
In this paper, we present calibrated preference optimization, a robust preference learning objective that fine-tunes the diffusion models to align with human preference by using multiple reward models. Specifically, we propose a simple, yet effective method to calibrate the rewards to approximate the win-rate against the base model. We then propose a diffusion preference optimization objective that regresses the difference between the calibrated rewards, which effectively learns from the reward without over-optimization. Furthermore, we extend our approach to a multi-reward problem by providing a frontier-based rejection sampling method that enables joint optimization of various reward signals. Extensive experimental results demonstrate that our approach is efficient and can boost the model performance without using any human-collected preference dataset.

\section*{Acknowledgements}
Kyungmin Lee acknowledges the partial support from 
Artificial Intelligence Graduate School Program (KAIST) (RS-2019-II190075) and
Institute of Information \& communications Technology Planning \& Evaluation (IITP) (No.RS-2021-II212068, Artificial Intelligence Innovation Hub).

{
    \small
    \bibliographystyle{ieeenat_fullname}
    \bibliography{main}
}

\clearpage

\onecolumn
\appendix
\begin{center}
    {\bf {\Large Calibrated Multi-Preference Optimization for Aligning Diffusion Models}} \\ 
    \vspace{3mm}
    { \Large Supplementary Materials}
    
\end{center}

\section{Additional description}\label{appendix:add_desc}
In this section, we provide additional details to Sec. 3 and Sec. 4 of the main manuscript. Specifically, we review the preliminaries on diffusion models and flow-based models (Sec.~\ref{sec:bg}), preference optimization for diffusion models (Sec.~\ref{sec:diffusionpo}), and provide details on loss weighting scheme (Sec.~\ref{sec:lossweighting}).

\subsection{Background on diffusion and flow-based models}\label{sec:bg}

\vspace{0.05in}
\noindent
{\bf Diffusion models.}
Let $q(\mathbf{x})$ be the density of data distribution of a sample $\mathbf{x}$ and $p_\theta(\mathbf{x})$ be a generative model parameterized by $\theta$ that approximates $q$. 
Given $\mathbf{x}\sim q(\mathbf{x})$, the diffusion model considers a series of latent variables $\mathbf{x}_t$ at time $t\in[0,1]$. 
Specifically, the forward process forms a conditional distribution $q(\mathbf{x}_t|\mathbf{x})$, where the marginal distribution is given by 
\begin{equation}
\mathbf{x}_t = \alpha_t \mathbf{x} + \sigma_t \boldsymbol{\epsilon}\text{,}
\end{equation}
where $\boldsymbol{\epsilon}\sim\mathcal{N}(\boldsymbol{0}, \mathbf{I})$, and $\alpha_t, \sigma_t$ are noise scheduling functions such that satisfies $\alpha_0 \approx 1$, $\alpha_1\approx 0$, and $\sigma_0 \approx 0$, $\sigma_1 \approx 1$.
Let us denote $\lambda_t = \log (\alpha_t^2 / \sigma_t^2)$ log signal-to-noise ratio (log-SNR), then $\lambda_t$ is a decreasing function of $t$.
Here, $\alpha_t$ and $\sigma_t$ (or equivalently $\lambda_t$) is chosen to satisfy that $\mathbf{x}_1$ is indiscernible from Gaussian noise (\emph{i.e.}, $p(\mathbf{x}_1)\approx \mathcal{N}(\boldsymbol{0}, \mathbf{I}))$, and conversely, $\mathbf{x}_0$ matches the data density $q(\mathbf{x})$. 
Then the reverse generative process gradually denoises the random Gaussian noise $\mathbf{x}_1\sim\mathcal{N}(\boldsymbol{0},\mathbf{I})$ to recover $\mathbf{x}_0$.
Specifically, the sampling process is governed by solving time-discretized SDE~\citep{song2020score, ho2020denoising} or probability flow ODE~\citep{song2020denoising, karras2022elucidating}, by using the score function $\nabla \log q(\mathbf{x}_t)$.
Training diffusion model then optimizes the neural network to approximate the score function by $\mathbf{s}_\theta(\mathbf{x}_t;t)$.
Especially, using the noise-prediction model~\citep{ho2020denoising} is a common practice, where the training objective can be written as following weighted loss objective~\citep{kingma2023understanding}:
\begin{equation}\label{eq:epsloss}
    \mathcal{L}_{\textrm{DM}}(\theta;\mathbf{x}) =\mathbb{E}_{t\sim \mathcal{U}(0,1), \boldsymbol{\epsilon}\sim\mathcal{N}(\boldsymbol{0},\mathbf{I})}\big[-\tfrac{1}{2}w_t\lambda_t'\|\boldsymbol{\epsilon}_\theta(\mathbf{x}_t;t) - \boldsymbol{\epsilon}\|_2^2\big]\text{,}
\end{equation}
where $w_t$ is a weighting function and $\lambda_t'$ is a time-derivative of $\lambda_t$.
Note that when $w_t=1$ for all $t\in(0,1)$, it becomes the variational lower bound (vlb) of KL divergence~\citep{kingma2021variational}, and the original DDPM uses $w_t\lambda_t'=-1$. 

\vspace{0.05in}
\noindent
{\bf Flow models.}
Alternatively, flow-based models or stochastic interpolants~\citep{lipman2022flow, albergo2023stochastic, ma2024sit} consider approximating the velocity field $\boldsymbol{v}(\mathbf{x}_t,t)$ on $\mathbf{x}$ at time $t\in(0,1)$, and solve following probability flow ODE to transport noise to data distribution:
\begin{equation}
    \mathbf{x}_t' = \boldsymbol{v}(\mathbf{x}_t, t)\text{,}
\end{equation}
where the marginal distribution of the solution of ODE matches the distribution $q_t(\mathbf{x}_t)$.
Given $\mathbf{x}_t=\alpha_t\mathbf{x}+\sigma_t\boldsymbol{\epsilon}$ for some $t\in(0,1)$ and $\boldsymbol{\epsilon}\sim\mathcal{N}(\boldsymbol{0}, \mathbf{I})$, the velocity field satisfies following:
\begin{equation}
    \boldsymbol{v}(\mathbf{x}_t, t) = \mathbb{E}[{\mathbf{x}}_t' \,|\, \mathbf{X}_t = \mathbf{x}_t] = \alpha_t'\, \mathbb{E}[\mathbf{x} \,|\, \mathbf{X}_t = \mathbf{x}_t] + \sigma_t'\,\mathbb{E}[\boldsymbol{\epsilon}\,|\, \mathbf{X}_t= \mathbf{x}_t]\text{,}
\end{equation}
and training objective for flow matching model is given as follows:
\begin{equation}\label{eq:vloss}
    \mathcal{L}_{\textrm{FM}}(\theta) = \mathbb{E}_{t\sim\mathcal{U}(0,1), \boldsymbol{\epsilon}\sim\mathcal{N}(\boldsymbol{0}, \mathbf{I})}\big[ \| \boldsymbol{v}_\theta(\mathbf{x}_t,t) - (\alpha_t' \mathbf{x} + \sigma_t'\boldsymbol{\epsilon})\|_2^2\big]\text{.}
\end{equation}
Note that Eq.~\eqref{eq:vloss} is a special case of Eq.~\eqref{eq:epsloss}, when $w_t=-\tfrac{1}{2}\lambda_t'\sigma_t^2$~\citep{kingma2023understanding, esser2024scaling}. 
In case of Rectified Flow~\citep{lipman2022flow}, we set $\alpha_t= 1-t$, $\sigma_t = t$, and $\lambda_t = 2 \log (\tfrac{1-t}{t})$, and the training objective of rectified flow model is given as follows:
\begin{equation}\label{eq:rfloss}
    \mathcal{L}_{\textrm{RF}}(\theta) = \mathbb{E}_{t, \eps}[\|\boldsymbol{v}_\theta(\mathbf{x}_t,t) - (\eps - \mathbf{x})\|_2^2]\text{.}
\end{equation}
For SD3-M~\citep{esser2024scaling}, we use Eq.~\eqref{eq:rfloss} to compute the loss.

\subsection{Diffusion preference optimization}\label{sec:diffusionpo}
For preference optimization with diffusion models, we consider following relaxation of original RLHF objective:
\begin{equation}\label{eq:diffobj}
    \max_\theta \bar{R}(\mathbf{x}_{0:1}, \mathbf{c}) - \beta D_{\textrm{KL}}\big(p_\theta(\mathbf{x}_{0:1}|\mathbf{c}) \,\|\, p_{\textrm{ref}}(\mathbf{x}_{0:1} | \mathbf{c})\big)\text{,}
\end{equation}
where $\bar{R}(\mathbf{x}_{0:1}, \mathbf{c})$ satisfies following:
\begin{equation}
    R(\mathbf{x}, \mathbf{c}) = \mathbb{E}_{q(\mathbf{x}_{0:1}|\mathbf{x})}
    \big[ \bar{R}(\mathbf{x}_{0:1}, \mathbf{c})\big] \text{.}
\end{equation}
Then by rearranging the equation derived from the closed solution of Eq.~\eqref{eq:diffobj}, we have following:
\begin{equation}\label{eq:diffreward}
    \bar{R}(\mathbf{x}_{0:1}, \mathbf{c}) = \beta \log\frac{p_\theta(\mathbf{x}_{0:1}|\mathbf{c})}{p_{\textrm{ref}}(\mathbf{x}_{0:1}|\mathbf{c})} - \beta \log Z(\mathbf{c})\text{,}
\end{equation}
where $Z(\mathbf{c})$ is a partition function. From Eq.~\eqref{eq:diffreward} and by rearranging $q(\mathbf{x}_{0:1} | \mathbf{x})$ in the inside term, we have
\begin{align}\label{eq:derivation1}
\begin{split}
    \mathbb{E}_{q(\mathbf{x}_{0:1}|\mathbf{x})}[\bar{R}(\mathbf{x}_{0:1},\mathbf{c})-\beta\log Z(\mathbf{c})] 
    &= \mathbb{E}_{q(\mathbf{x}_{0:1}|\mathbf{x})}\bigg[\beta\log\frac{p_\theta(\mathbf{x}_{0:1}|\mathbf{c}) }{q(\mathbf{x}_{0:1}|\mathbf{x})} - \beta\log \frac{p_{\textrm{ref}}(\mathbf{x}_{0:1}|\mathbf{c})}{q(\mathbf{x}_{0:1}|\mathbf{x})}\bigg] \\
    &= \beta \big(D_{\textrm{KL}}(q(\mathbf{x}_{0:1}|\mathbf{x}) \,\|\, p_{\textrm{ref}}(\mathbf{x}_{0:1}|\mathbf{c})) - D_{\textrm{KL}}(q(\mathbf{x}_{0:1}|\mathbf{x}) \,\|\, p_\theta(\mathbf{x}_{0:1}|\mathbf{c}))\big)\text{.}
\end{split}
\end{align}
Note that the KL divergence satisfies following (see \citep{kingma2023understanding} for details):
\begin{equation}\label{eq:diffeps}
    \frac{\mathrm{d}}{\mathrm{d}t}D_{\textrm{KL}}\big(q(\mathbf{x}_{t:1}|\mathbf{x}\,\|\, p_\theta(\mathbf{x}_{t:1}|\mathbf{c})\big) =  \frac{1}{2}\lambda_t'\mathbb{E}_{\boldsymbol{\epsilon}\sim\mathcal{N}(\boldsymbol{0}, \mathbf{I})}\big[ \|\boldsymbol{\epsilon}_\theta(\mathbf{x}_t;\mathbf{c},t) - \boldsymbol{\epsilon}\|_2^2 \big]\text{.} 
\end{equation}
By taking integration of Eq.~\eqref{eq:diffeps} over $t\in (1,0)$, one can rewrite $R(\mathbf{x}, \mathbf{c})$ as follows:
\begin{equation}
    R(\mathbf{x}, \mathbf{c}) = \frac{\beta}{2}\mathbb{E}_{t\sim\mathcal{U}(0,1), \boldsymbol{\epsilon}\sim\mathcal{N}(0,1)}\big[ \lambda_t'\big(\|\boldsymbol{\epsilon}_\theta(\mathbf{x}_t;\mathbf{c},t) - \boldsymbol{\epsilon}\|_2^2 - \|\boldsymbol{\epsilon}_\phi(\mathbf{x}_t;\mathbf{c},t) - \boldsymbol{\epsilon}\|_2^2 \big)\big]\text{,}    
\end{equation}
For a triplet $(\mathbf{c}, \mathbf{x}^+, \mathbf{x}^-)$, we consider following upper bound of a training objective for any convex function $g:\mathbb{R}\rightarrow\mathbb{R}$:
\begin{align*}
\begin{split}
    \bar{\ell}(\theta) &= g\big(R(\mathbf{x}^+, \mathbf{c}) - R(\mathbf{x}^-, \mathbf{c})\big) \\
    &= g\bigg( \frac{\beta}{2}\mathbb{E}_{t, \boldsymbol{\epsilon}^+, \boldsymbol{\epsilon}^-}\big[ \lambda_t'\big(\|\boldsymbol{\epsilon}_\theta(\mathbf{x}_t^+;\mathbf{c},t) - \boldsymbol{\epsilon}^+\|_2^2 - \|\boldsymbol{\epsilon}_{\textrm{ref}}(\mathbf{x}_t^+;\mathbf{c},t) - \boldsymbol{\epsilon}^+\|_2^2 - \|\boldsymbol{\epsilon}_\theta(\mathbf{x}_t^-;\mathbf{c},t) - \boldsymbol{\epsilon}^-\|_2^2 + \|\boldsymbol{\epsilon}_{\textrm{ref}}(\mathbf{x}_t^-;\mathbf{c},t) - \boldsymbol{\epsilon}^-\|_2^2 \big)\big] \bigg)\\
    &\leq \mathbb{E}_{t,\boldsymbol{\epsilon}^+, \boldsymbol{\epsilon}^-}\bigg[g\bigg(
    \tfrac{1}{2}\beta\lambda_t'\big(\|\boldsymbol{\epsilon}_\theta(\mathbf{x}_t^+;\mathbf{c},t) - \boldsymbol{\epsilon}^+\|_2^2 - \|\boldsymbol{\epsilon}_{\textrm{ref}}(\mathbf{x}_t^+;\mathbf{c},t) - \boldsymbol{\epsilon}^+\|_2^2 - \|\boldsymbol{\epsilon}_\theta(\mathbf{x}_t^-;\mathbf{c},t) - \boldsymbol{\epsilon}^-\|_2^2 + \|\boldsymbol{\epsilon}_{\textrm{ref}}(\mathbf{x}_t^-;\mathbf{c},t) - \boldsymbol{\epsilon}^-\|_2^2\big)\bigg)
    \bigg]\text{,}
\end{split}
\end{align*}
where $t\sim\mathcal{U}(0,1)$, $\boldsymbol{\epsilon}^+\sim\mathcal{N}(\boldsymbol{0}, \mathbf{I})$, $\boldsymbol{\epsilon}^- \sim \mathcal{N}(\boldsymbol{0}, \mathbf{I})$, and the last inequality comes from the Jensen's inequality.
Using the equation we defined in our main paper, \emph{i.e.},
\begin{equation}\label{eq:diffratio2}
    R_\theta(\mathbf{x}_t, \mathbf{c}, t)  \! = \!  \lambda_t'\big(\|\boldsymbol{\epsilon}_\theta(\mathbf{x}_t;\mathbf{c},t) - \boldsymbol{\epsilon}\|_2^2 - \|\boldsymbol{\epsilon}_{\textrm{ref}}(\mathbf{x}_t;\mathbf{c},t) -\boldsymbol{\epsilon}\|_2^2\big)\text{,}
\end{equation}
we derive following training objectives for DPO, IPO, and CaPO:
\begin{align}
\begin{split}
    \ell_{\textrm{DPO}}(\theta) &= \mathbb{E}_{t,\boldsymbol{\epsilon}^+, \boldsymbol{\epsilon}^-}\bigg[ -\log \sigma \big(\beta \big(R_\theta(\mathbf{x}^+, \mathbf{c}, t) - R_\theta(\mathbf{x}^-, \mathbf{c}, t )\big)\big)\bigg]\\
    \ell_{\textrm{IPO}}(\theta) &= \mathbb{E}_{t,\boldsymbol{\epsilon}^+, \boldsymbol{\epsilon}^-}\bigg[ \bigg(1 -  \beta\big(R_\theta(\mathbf{x}^+, \mathbf{c}, t) - R_\theta(\mathbf{x}^-, \mathbf{c}, t )\big)\bigg)^2\bigg]\\
    \ell_{\textrm{CaPO}}(\theta) &= \mathbb{E}_{t,\boldsymbol{\epsilon}^+, \boldsymbol{\epsilon}^-}\bigg[ \bigg(R(\mathbf{x}^+, \mathbf{c}) - R(\mathbf{x}^-, \mathbf{c}) -  \beta\big(R_\theta(\mathbf{x}^+, \mathbf{c}, t) - R_\theta(\mathbf{x}^-, \mathbf{c}, t )\big)\bigg)^2\bigg]\text{,}
\end{split}
\end{align}
where $R(\mathbf{x}, \mathbf{c})$ is a reward from the external reward model.

\vspace{0.05in}
\noindent
{\bf Independent noise sampling.}
Note that in original Diffusion-DPO paper~\citep{wallace2023diffusion}, the author proposed to use same noise for $\mathbf{x}^+$ and $\mathbf{x}^-$, \emph{i.e.},  $\boldsymbol{\epsilon}^+ = \boldsymbol{\epsilon}^-$, while we sample independent noise for $\eps^+$ and $\eps^-$. We believe this is more theoretically grounded, and empirically found that it has slightly better performance than using the same noise (even for DPO and IPO).

\begin{figure}[t]
    \centering
    \begin{subfigure}[t]{0.35\columnwidth}
        \includegraphics[width=\linewidth]{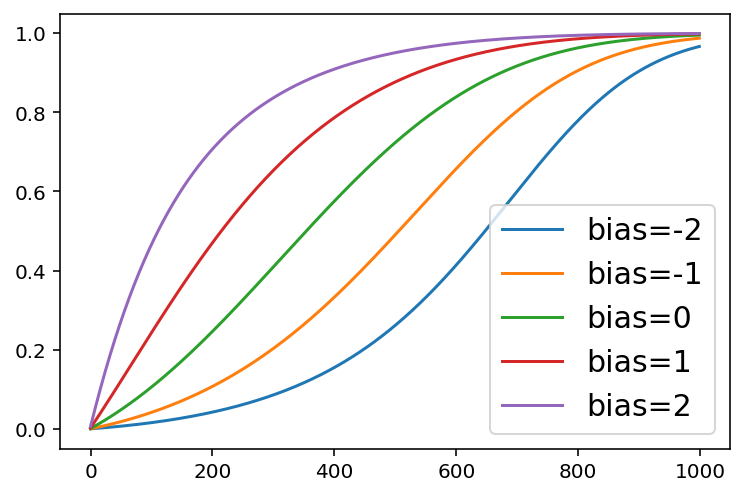}
        \caption{Weighting for noise-prediction loss}
        \label{fig:eps_weight}
    \end{subfigure}
    \begin{subfigure}[t]{0.35\columnwidth}
        \centering\small
        \includegraphics[width=\linewidth]{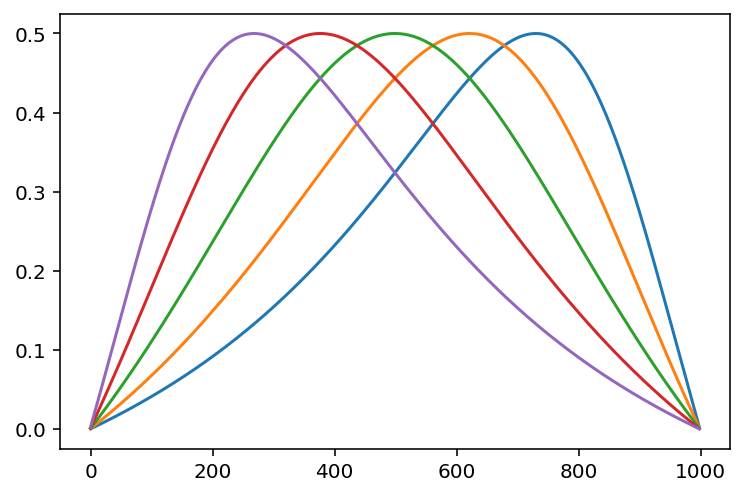}
        \caption{Weighting for flow matching loss}
        \label{fig:v_weight}
    \end{subfigure}
    \caption{\textbf{Loss weighting.} We plot the weighting function with bias $b\in \{-2, -1, 0, 1, 2\}$ for each noise prediction loss and flow matching loss.} 
    \label{fig:loss_weight}
    \vspace{-5pt}
\end{figure}
\subsection{Loss weighting}\label{sec:lossweighting}
In practice, we multiply $w_t$ to the noise-prediction loss for diffusion preference optimization. One can consider this as setting timestep-wise different $\beta_t = \beta w_t$, \emph{i.e.}, giving different regularization hyperparameters at each time $t\in(0,1)$. Thus, we have 
\begin{equation}\label{eq:diffratioweight}
    R_\theta(\mathbf{x}_t, \mathbf{c}, t)  \! = \!  w_t\lambda_t'\big(\|\boldsymbol{\epsilon}_\theta(\mathbf{x}_t;\mathbf{c},t) - \boldsymbol{\epsilon}\|_2^2 - \|\boldsymbol{\epsilon}_{\textrm{ref}}(\mathbf{x}_t;\mathbf{c},t) -\boldsymbol{\epsilon}\|_2^2\big)\text{,}
\end{equation}
and applies to each DPO, IPO, and CaPO loss. 
As we mentioned in Sec.~4.4 in our main draft, we use sigmoid loss weighting~\citep{kingma2023understanding}, where the loss weights are sigmoid function of log-SNR $\lambda_t$ with bias $b$:
\begin{equation}
    w_t = w(\lambda_t) = \frac{1}{1+\exp(b - \lambda_t)}\text{.}
\end{equation}
Note that SDXL uses a modified DDPM schedule~\citep{ho2020denoising}, where $\beta = \big(\sqrt{\beta_0} + \tfrac{t}{T-1}(\sqrt{\beta_{T-1}} - \sqrt{\beta_0})\big)^2$, and $\alpha_t = (\prod_{s=0}^t (1-\beta_s))^{1/2}$.
Since it is impractical to compute $\lambda_t'$, we simply set it as constant (\emph{i.e.}, linear $\lambda_t$, which empirically holds when $\lambda_t \in [-15, 5]$, and for $\lambda_t > 0.5$ the weight $w_t$ is close to $0$, so one can ignore it).

SD3-M uses a rectified flow scheduler~\citep{lipman2022flow}, where $\lambda_t = 2 \log (\tfrac{1-t}{t})$. Note that we have 
\begin{equation}
\mathbb{E}_{t\sim\mathcal{U}(0,1),\eps}\big[\|\boldsymbol{v}_\theta(\mathbf{x}_t,t) - (\eps - \mathbf{x})\|_2^2\big] = \mathbb{E}_{\lambda\sim \mathcal{U}(\lambda_{\textrm{min}}, \lambda_{\textrm{max}}), \eps}\big[e^{-\lambda/2}\|\eps_\theta(\mathbf{x}_\lambda,\lambda) - \eps\|_2^2\big]\text{,}
\end{equation}
where $\mathbf{x}_\lambda$ denotes forward process of $\mathbf{x}$ with log-SNR value $\lambda$, and $\lambda_{\textrm{min}}$ and $\lambda_{\textrm{max}}$ denotes the minimal and maximal value for log-SNR (see \citep{kingma2023understanding} Appendix D.3 for details).
As such, multiplying $w = \sigma(-\lambda)$ to noise-prediction loss is equivalent to multiplying $(e^{\lambda/2} + e^{-\lambda / 2})^{-1}$ to flow matching objective:
\begin{equation}
    \sigma(-\lambda)\|\eps_\theta(\mathbf{x}_\lambda,\lambda) - \eps\|_2^2 = \sigma(-\lambda)\cdot \frac{e^{-\lambda / 2}}{e^{-\lambda / 2}}\|\eps_\theta(\mathbf{x}_\lambda,\lambda) - \eps\|_2^2 = \frac{1}{e^{\lambda / 2} + e^{-\lambda / 2}}\|\boldsymbol{v}_\theta(\mathbf{x}_\lambda, \lambda) - (\eps - \mathbf{x})\|_2^2\text{.} 
\end{equation}
If we shift with bias $b$, it becomes $w_\lambda = (e^{-(\lambda -b)/2} + e^{(\lambda -b)/2})^{-1}$.
In Fig.~\ref{fig:loss_weight}, we plot loss weighting functions for noise-prediction loss and flow matching loss with different bias $b\in\{-2, -1, 0, 1, 2\}$.
In practice, we select $\lambda \sim \mathcal{U}[-10, 10]$ and multiply $w_\lambda$ to flow matching loss.
Note that multiplying $w_\lambda$ has a similar effect in log-normal sampling proposed in \citep{karras2022elucidating, esser2024scaling}, where we empirically find similar performance. To ensure consistency with SDXL experiments, we use loss weighting instead of logit-normal sampling for SD3-M experiments.

\begin{figure}[t]
    \centering
    \begin{subfigure}[t]{0.32\columnwidth}
        \includegraphics[width=\linewidth]{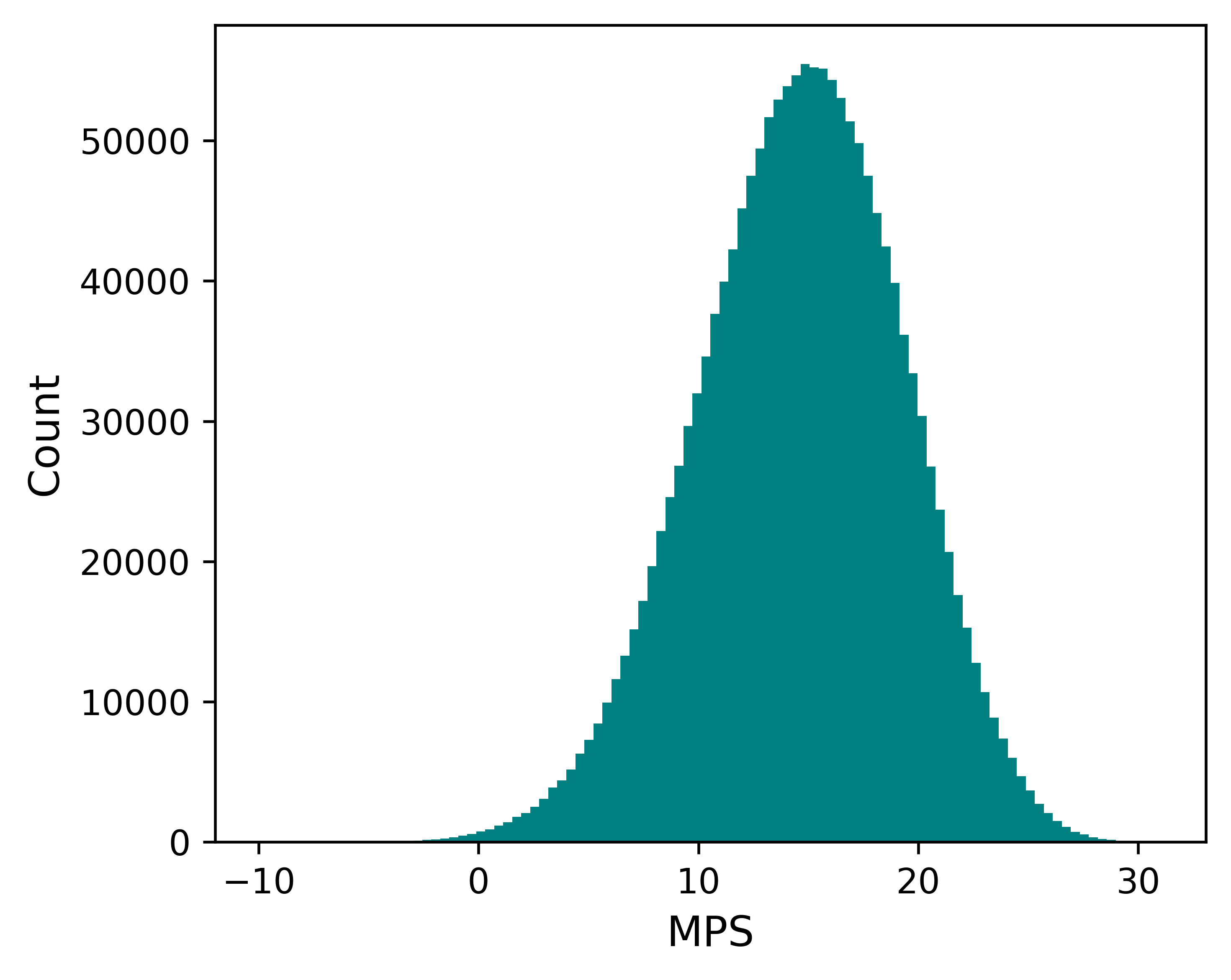}
        \caption{MPS scores of training dataset}
        \label{fig:mps_train}
    \end{subfigure}
    \hfill
    \begin{subfigure}[t]{0.32\columnwidth}
        \centering
        \includegraphics[width=\linewidth]{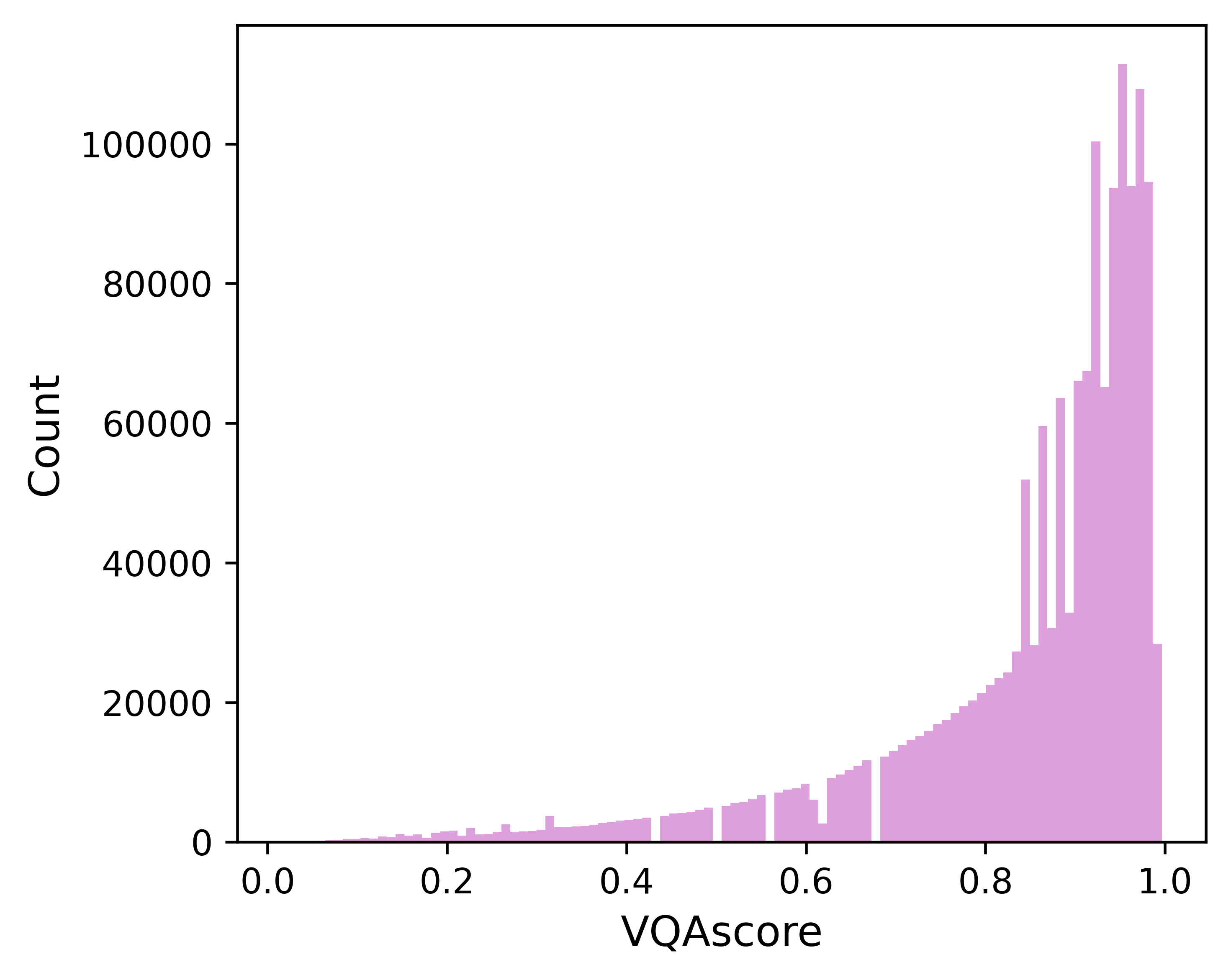}
        \caption{VQAscores of training dataset}
        \label{fig:vqa_train}
    \end{subfigure}
    \hfill
    \begin{subfigure}[t]{0.32\columnwidth}
        \centering\small
        \includegraphics[width=\linewidth]{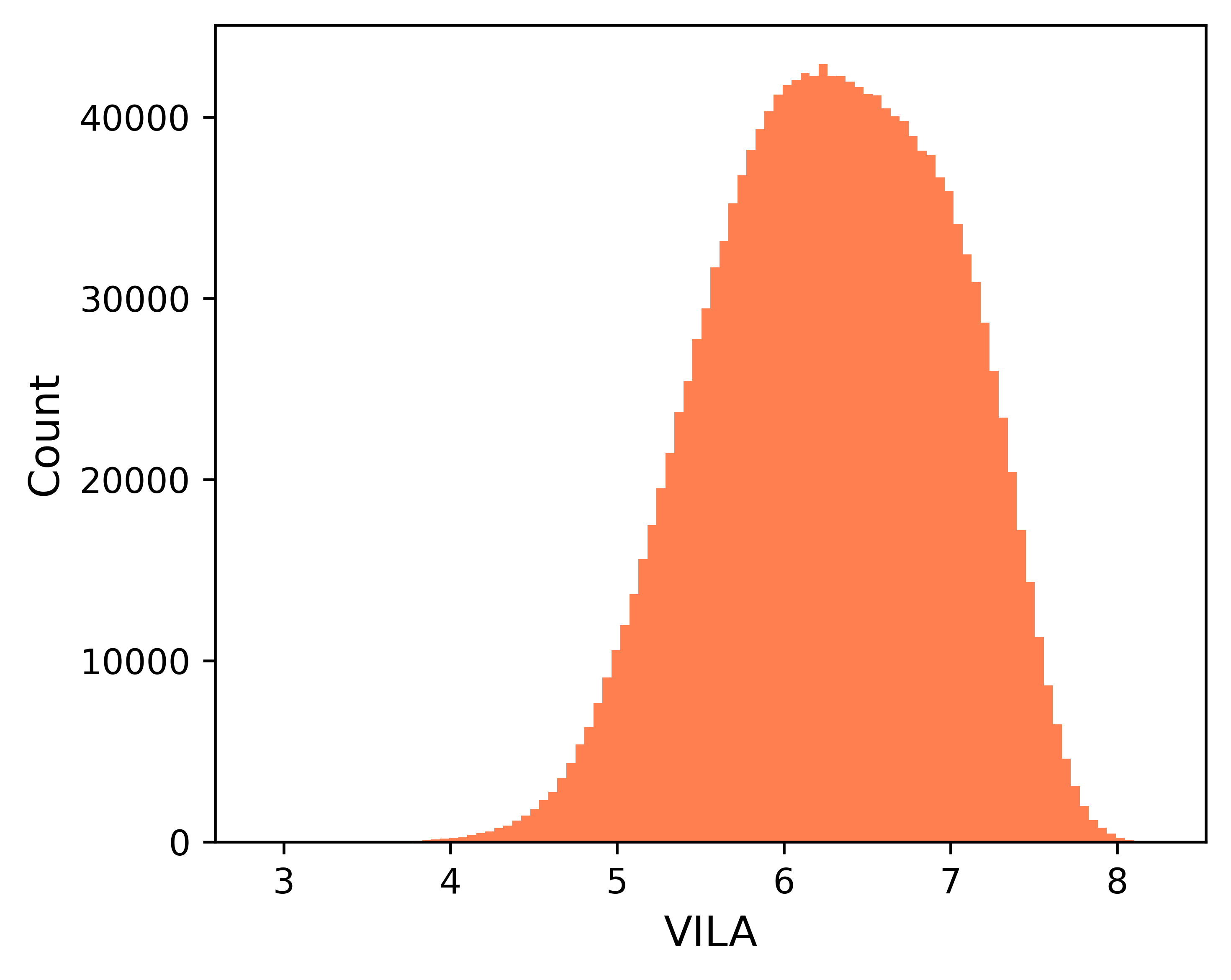}
        \caption{VILA scores of training dataset}
        \label{fig:vila_train}
    \end{subfigure}
    \vfill
    \begin{subfigure}[t]{0.32\columnwidth}
        \includegraphics[width=\linewidth]{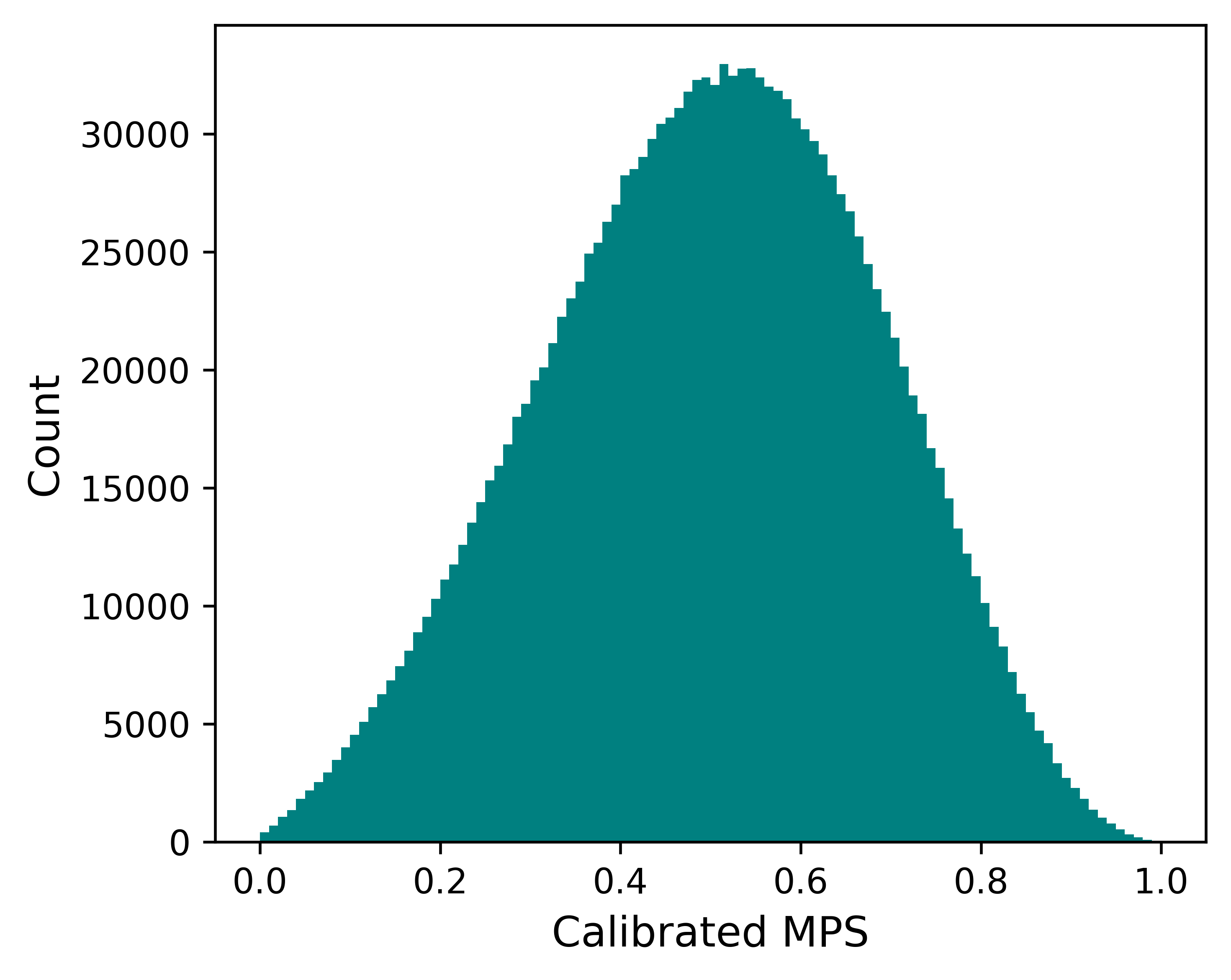}
        \caption{Calibrated MPS scores of training dataset}
        \label{fig:mps_cal_train}
    \end{subfigure}
    \hfill
    \begin{subfigure}[t]{0.32\columnwidth}
        \centering
        \includegraphics[width=\linewidth]{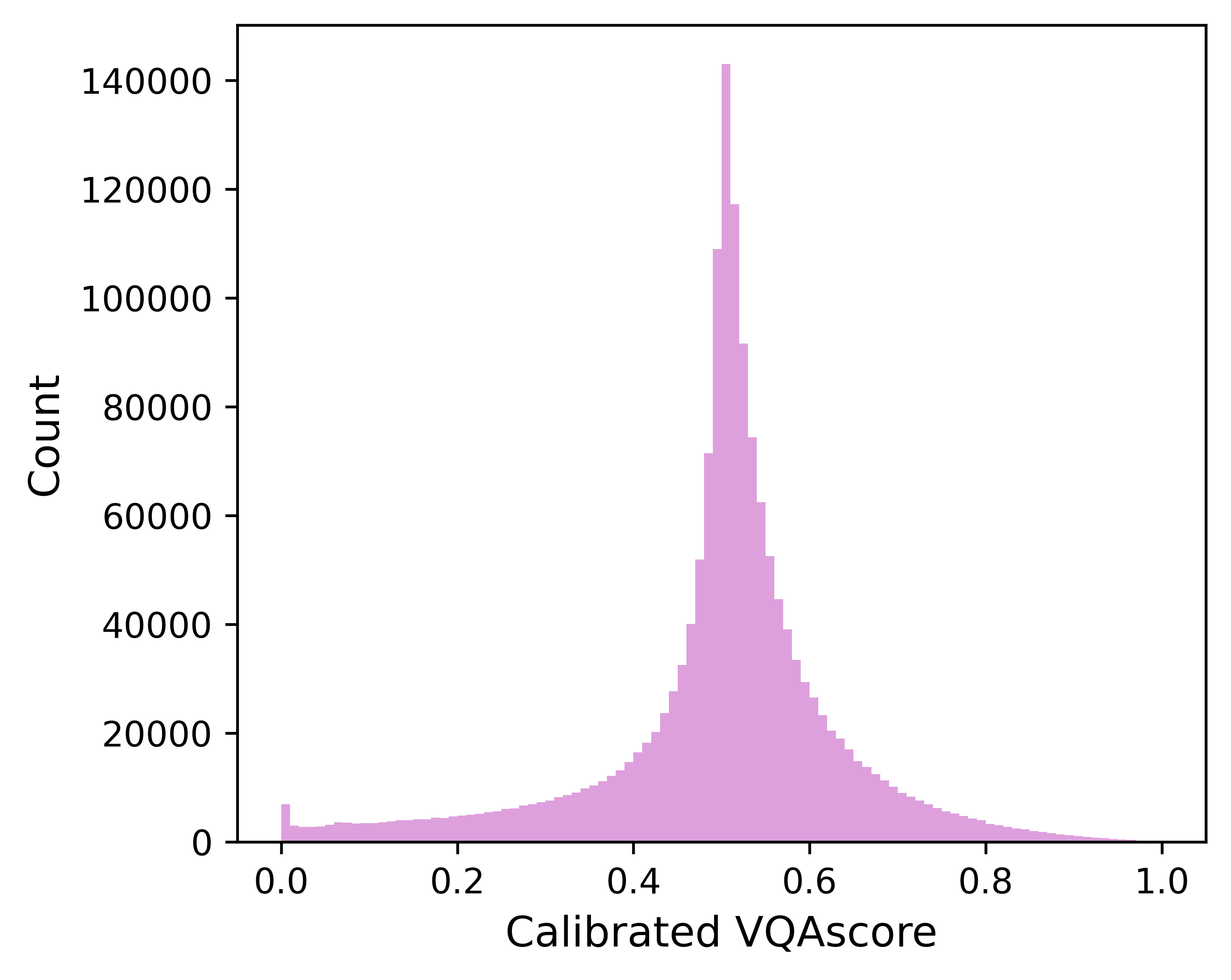}
        \caption{Calibrated VQAscores of training dataset}
        \label{fig:vqa_cal_train}
    \end{subfigure}
    \hfill
    \begin{subfigure}[t]{0.32\columnwidth}
        \centering\small
        \includegraphics[width=\linewidth]{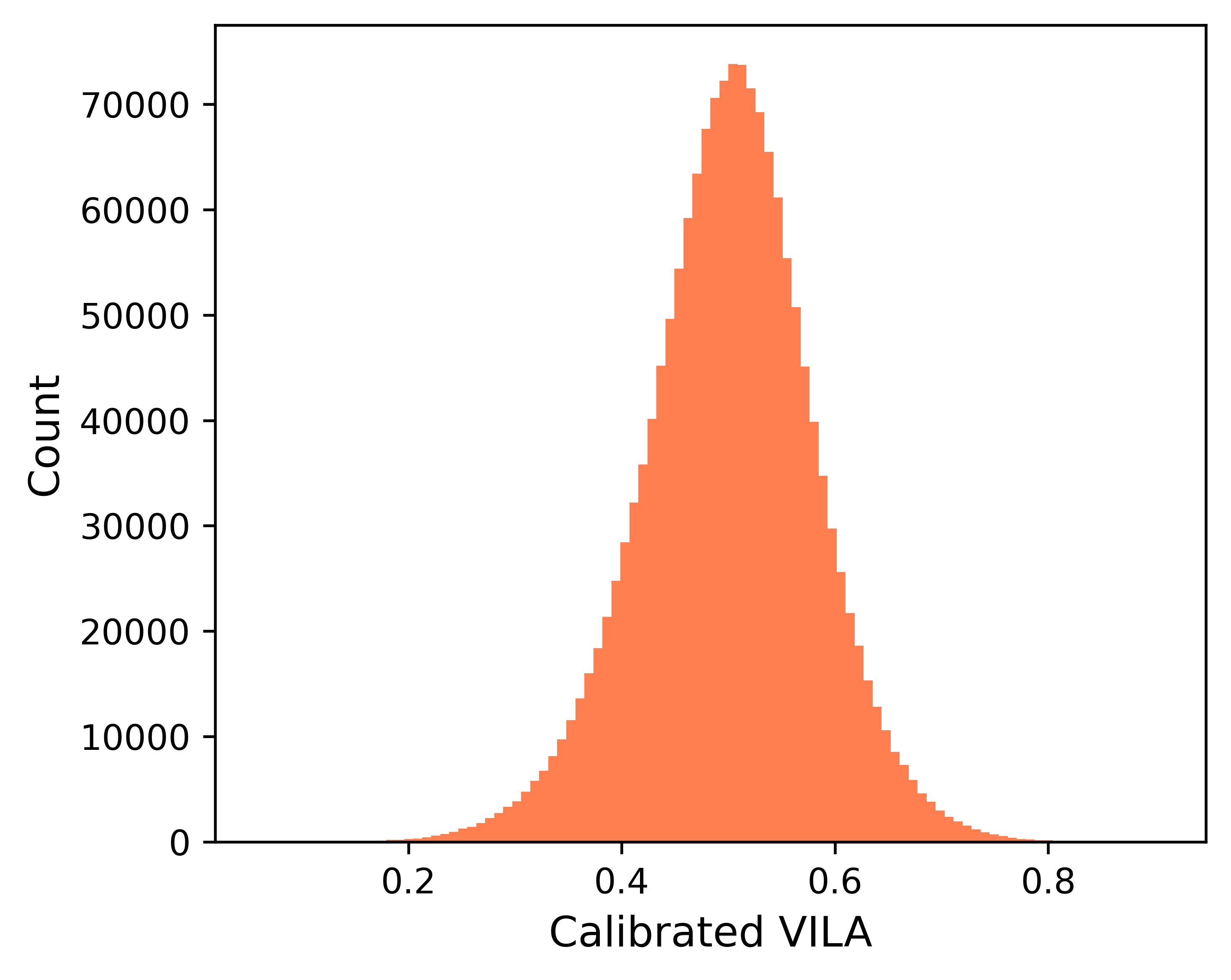}
        \caption{Calibrated VILA scores of training dataset}
        \label{fig:vila_cal_train}
    \end{subfigure}
    \vspace{-5pt}
    \caption{\textbf{Training dataset score distribution.} We plot the histogram of rewards (top row) and calibrated rewards (bottom row) of training dataset. By using calibration, the scores are centered and bounded in range $[0,1]$.} 
    \label{fig:dataset_statistic}
    \vspace{-10pt}
\end{figure}
\section{Implementation Details}
\subsection{Dataset}\label{appendix:dataset}
\vspace{0.05in}
\noindent
{\bf Reward models.}
For reward models learned by fine-tuning CLIP models (\emph{e.g.}, Pickscore~\citep{kirstain2023pick}, MPS~\citep{zhang2024learning}), we compute the reward by the dot product between the image embedding and the text embedding. To compute MPS score, we additionally multiply the text embedding from condition textual description (\emph{e.g.}, textual description for aesthetic quality).
For VQAscore, we use CLIP-FlanT5-XXL~\citep{lin2025evaluating}, and compute the score by probability of {\tt{"Yes"}} token given the image and question provided to the model: 
\begin{equation}
    P\big({\texttt{"Yes"}} |\mathbf{x}, {\texttt{"Does this figure shows \{prompt\}? Please answer yes or no."}}\big)\text{,}
\end{equation}
where $\mathbf{x}$, {\tt{prompt}} are image and text input.
While VQAscore is not a Bradley-Terry model, we simply approximate the win-rate by following:
\begin{equation}
    \mathbb{P}(\mathbf{x} \succ \mathbf{x}' | \mathbf{c}) =\frac{ s(\mathbf{x}, \mathbf{c})^\alpha}{ s(\mathbf{x}, \mathbf{c})^\alpha + s(\mathbf{x}', \mathbf{c})^\alpha}\text{,}
\end{equation}
where $s(\mathbf{x}, \mathbf{c})$ is a VQAscore and $\alpha > 0$ is a hyperparameter to control the temperature. We find $\alpha=1$ works well in our experiments.
Lastly, VILA-R score~\citep{ke2023vila} outputs the aesthetic score between 1-10, and we apply the Bradley-Terry model to compute the win-rate.
In Fig.~\ref{fig:dataset_statistic}, we plot the histogram of reward scores and calibrated rewards of our training dataset.

\vspace{0.05in}
\noindent
{\bf Training dataset.}
We use 100K prompts from DiffusionDB~\citep{wang2022diffusiondb} and generate $N=16$ images per prompt.
For SDXL, we use DDIM~\citep{song2020denoising} scheduler, guidance scale of 7.5 and sampling steps of 50. 
For SD3-M, we use the DPM solver~\citep{lu2022dpm} for flow-based models, guidance scale of 5.0 and sampling steps of 50. 
Furthermore, as described in \citep{esser2024scaling}, we shift the timestep schedules to reside more on higher timesteps, \emph{i.e.}, we set $t \leftarrow \frac{ts}{1+ t(s-1)}$ with shift scale $s=3.0$.
 
\subsection{Training and evaluation}
\vspace{0.05in}
\noindent
{\bf Training configuration.} Throughout experiments, we use Jax~\citep{frostig2018compiling} and train models using the Optax library on TPU chips.
For both SDXL and SD3-M experiments, we use Adam~\citep{kingma2014adam} optimizer.
Regarding training configuration for SDXL experiments, we use batch size of 256, learning rate of 1e-5 with linear warmup for first 1000 steps, and train for maximum 10000 steps. 
For SD3-M, we use batch size of 256, learning rate of 1.5e-5 with linear warmup for first 1000steps, and train for maximum 5000 steps.
We choose hyperparameter $\beta$ by sweeping over $\{300, 500, 1000\}$ for CaPO, $\{500, 1000, 2000\}$ for IPO, and $\{2000, 3000, 4000\}$ for DPO when training SDXL model. 
For SD3-M, we sweep over $\{30, 50, 100\}$ for CaPO, $\{50, 100, 200\}$ for IPO, and $\{100, 200, 300\}$ for DPO.
For all training, we use sigmoid loss weighting with $b=1.5$ for SDXL and $b=-1.0$ for SD3-M (including all DPO, IPO, and CaPO). 
During training, we generate images using subset of Parti prompts at each 1000-th iteration, and choose the final model with maximum validation win-rate (average of win-rates for multi-reward signals).

\begin{figure}[t]
    \small\centering
    \includegraphics[width=0.70\columnwidth]{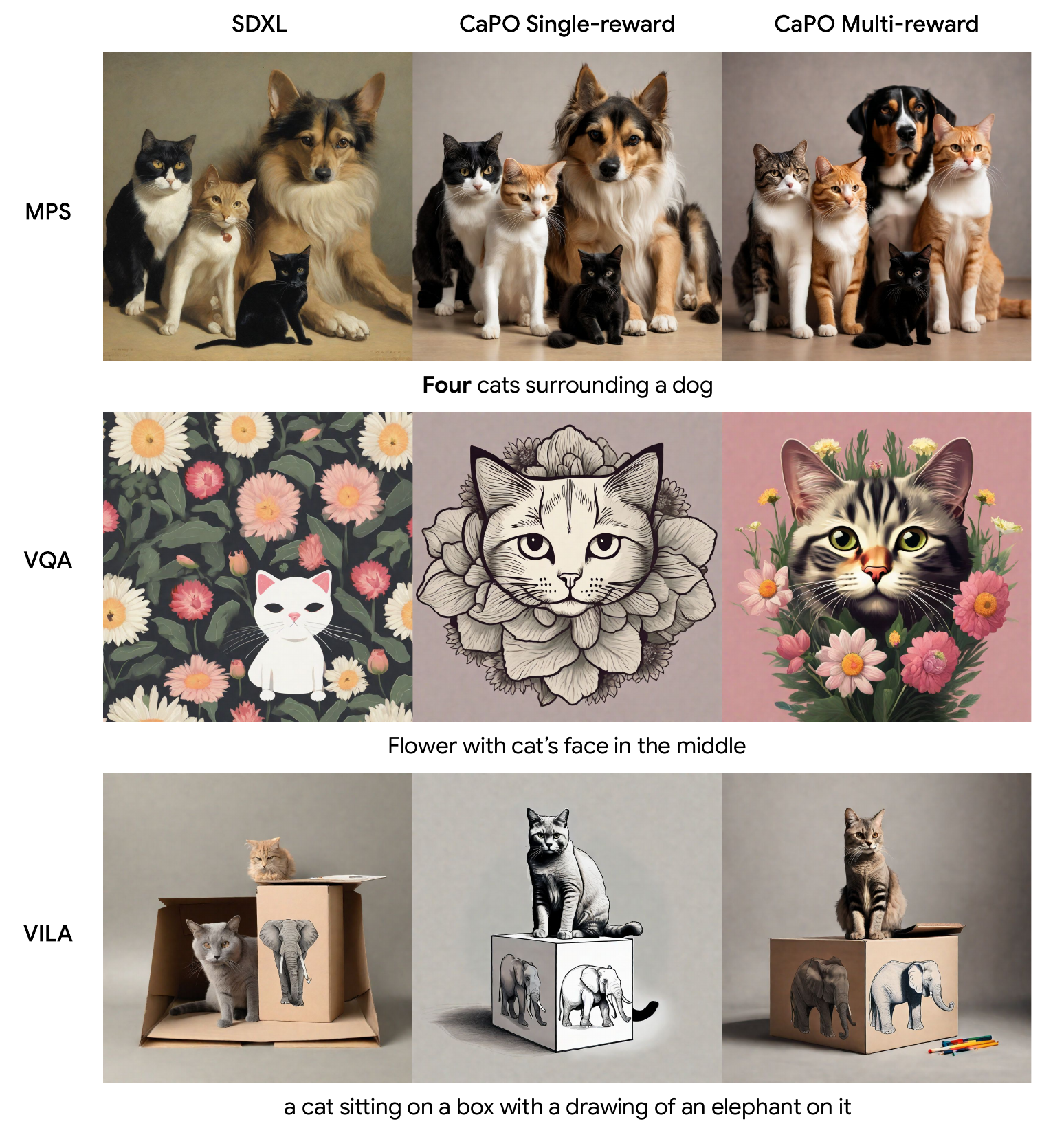}
    \caption{
    \textbf{Effect of multi-reward CaPO}. We demonstrate the qualitative comparison between CaPO with single-reward with each MPS, VQAscore, and VILA, and CaPO with multi-reward on SDXL.
    We see that optimizing with single-reward improves upon the base model, yet multi-reward CaPO shows the best overall improvement.
    For example, while using VQAscore alone improves the image-text alignment, the image aesthetics are significantly improved when using multi-reward. 
    Also, when using VILA score, the image aesthetics improve, but it often lose the image-text alignment (\emph{e.g.}, the image becomes drawing style, while only the elephant should be in a drawing style).}
    \label{fig:multi_effect}
\end{figure}

\vspace{0.05in}
\noindent
{\bf Model soup.}
For model merging experiments, we follow \citep{rame2024warp}. Specifically, suppose $\theta_0$ be weights of a pretrained model and $\theta_1$, $\theta_2$ be weights of fine-tuned models. Then the spherical linear interpolation (SLERP) between $\theta_1$ and $\theta_2$ is given by 
\begin{equation}
    \textrm{SLERP}(\theta_0, \theta_1, \theta_2, \lambda) = \theta_0 + \frac{\sin ((1-\lambda)\Omega)}{\sin (\Omega)} (\theta_1 - \theta_0) + \frac{\sin (\lambda \Omega)}{\sin (\Omega)}(\theta_2 - \theta_0)\text{,}
\end{equation}
where $\Omega$ is the angle between two task vectors $\theta_1 - \theta_0$ and $\theta_2 - \theta_0$, and $\lambda \in (0,1)$ is a coefficient. 
To merge three fine-tuned models, we first merge two models with $\lambda=0.5$ to obtain $\theta_{12} = \textrm{SLERP}(\theta_0, \theta_1, \theta_2, 0.5)$, then merge $\theta_{12}$ and $\theta_3$ with $\lambda = 1/3$ to obtain final model.

\vspace{0.05in}
\noindent
{\bf Evaluation.}
For evaluation, we generate images with the same configuration as in Sec.~\ref{appendix:dataset} for different benchmark prompt dataset.
For SDXL, we use Parti~\citep{yu2022scaling} prompts, and for SD3-M we use DPG-bench~\citep{hu2024ella}. 
Then we compute the win-rate against the base model by comparing one-by-one comparison for each image, \emph{e.g.}, if we have $K$ images from base model and $K$ images from fine-tuned model, we make $K^2$ comparison and count the number of win and divide by $K^2$.
We also report the average reward scores.
Remark that the average reward scores and win-rate could show different trends, as the model achieves a higher score gain for some prompts, but it fails to improve on others. Thus, we found win-rate is a more general metric to see the generalization over different prompts.

\vspace{0.05in}
\noindent
{\bf Benchmark evaluation.}
We use GenEval~\citep{ghosh2024geneval} and T2I-Compbench~\citep{huang2023t2i} to evaluate our models. 
For T2I-Compbench, we use BLIP-VQA model~\citep{li2022blip} to evaluate Color, Shape, Texture, Complex, and UniDet~\citep{zhou2022simple} for Spatial, and CLIP~\citep{radford2021learning} for Non Spatial. 
For baselines, we compare with the state-of-the-art open-source text-to-image diffusion models Flux-dev (12B)~\citep{flux2024},  Flux-schnell (12B)~\citep{flux2024}, and Stable Diffusion 3.5-Large (8B)~\citep{esser2024scaling}. 
Since those models are much larger than SDXL (2.6B) and SD3 (2B), we remark that this is not a fair comparison, yet we show the comparable performance of our method.

\section{Additional ablation study}\label{appendix:impl}

\vspace{0.05in}
\noindent
{\bf Effect of multi-reward.}
We demonstrate the effect of multi-reward CaPO compared to single-reward CaPO. As we demonstrated in Tab.~1 and Tab.~2 in our main draft, the single-reward model achieves the best score in which they have trained with, but the other metrics score below the multi-reward cases. We showcase the qualitative examples on the effect of multi-reward preference optimization compared to single-reward cases in Fig.~\ref{fig:multi_effect}. We notice that single-reward fine-tuning is often imperfect, \emph{e.g.}, fine-tuning with only VILA score loses image-text alignment, and fine-tuning with only VQAscore lacks image aesthetics. On the other hand, multi-reward fine-tuning complements those issues and improves the overall image quality.

\vspace{0.05in}
\noindent
{\bf Loss weighting for SD3-M.}
We show the effect of loss weighting when training SD3-M models. Similar to SDXL, we compare the results of CaPO multi-reward fine-tuning with different bias parameters. In Tab.~\ref{tab:abl_appendix}, we show that sigmoid weighting with bias $b=-1.0$ shows the best result, outperforming the constant weighting counterpart. Note that for SDXL, $b=1.5$ performs the best, while for SD3-M, negative bias $b=-1.0$ performs the best. Remark that as SD3-M performs diffusion modeling on $16\times 128 \times 128$, and SDXL performs on $4\times 128\times 128$, the bias shifts toward negative as the total variance becomes higher, and the log-SNR should be increased~\citep{hoogeboom2023simple, hoogeboom2024simpler}.

\vspace{0.2in}
\noindent
{\bf User study evaluation.}
We conduct additional user evaluations to compare our method with base models. For SDXL vs CaPO+SDXL, we randomly select 200 prompts from Parti prompt dataset~\citep{yu2022scaling}, and for SD3-M vs CaPO+SD3-M, we randomly select 200 prompts from GenAI bench prompt dataset~\citep{li2024genai}. Additionally, we compare CaPO+SDXL with Diffusion-DPO~\citep{wallace2023diffusion} again with 200 randomly selected prompts. We give following instructions to the raters:
\begin{itemize}[leftmargin=1cm]
    \item Instruction: Given the text below, pick the left or the right image with better looking.
    \item Good example: Images are beautiful and following text description.
    \item Bad example: Images are not looking good or not following text description.
\end{itemize}
We use Amazon mechanical Turk~\citep{amt} and 5 raters answered to each pair. 
In Tab.~\ref{tab:user}, we show the results of user study. 
We observe that CaPO+SD3-M and CaPO+SDXL consistently outperform SD3-M and SDXL, respectively. Also, CaPO+SDXL outperforms Diffusion-DPO, yet the margin is smaller than CaPO+SDXL vs SDXL.

\begin{table}[t]
\centering
\small
\centering\small
\begin{tabular}{l  ccc}
\toprule
 & MPS & VQA & VILA \\
\midrule
Constant weighting  & 54.9  & 53.6 & 55.7  \\
\midrule
Sigmoid weighting ($b=0$)  & 57.0 & 53.9 & 66.9   \\
Sigmoid weighting ($b=-1.0$)  & 59.0 & 55.7 & 69.3   \\
\bottomrule
\end{tabular}
\caption{
\textbf{Ablation on loss weighting for SD3-M.} We show the results of CaPO multi-reward fine-tuning SD3-M with constant weighting (\emph{i.e.}, $-w_t\lambda_t'=1$), and sigmoid weighting by varying bias $b=0.0, -1.0$. Similar to SDXL, using sigmoid weighting shows better results than constant weighting, and $b=-1.0$ performs the best.
}\label{tab:abl_appendix}
\end{table}
\begin{table}[t]
\centering\small
\setlength\tabcolsep{2pt} 
\begin{tabular}{ ccc | ccc | ccc }
\toprule
 CaPO+SDXL & SDXL & Improvement & CaPO+SDXL & Diffuion-DPO & Improvement & CaPO+SD3-M & SD3-M & Improvement\\
\midrule
{ 54.5\%} & 45.5\% & {\bf +10\%} & 52.0\% & 48.0\% & {\bf +4\%} & { 53.5\%} & 46.5\% & {\bf +7\%}\\
\bottomrule
\end{tabular}
\caption{
\textbf{User study results.} We report the win-rate from the user study by using 200 images. We compare CaPO+SD3-M vs SD3-M, CaPO+SDXL vs SDXL, and CaPO+SDXL vs Diffusion-DPO~\citep{wallace2023diffusion}.
CaPO achieves consistent win against baseline. 
}\label{tab:user}
\end{table}

\newpage
\begin{figure}[!htb]
    \small\centering
    \includegraphics[width=0.98\columnwidth]{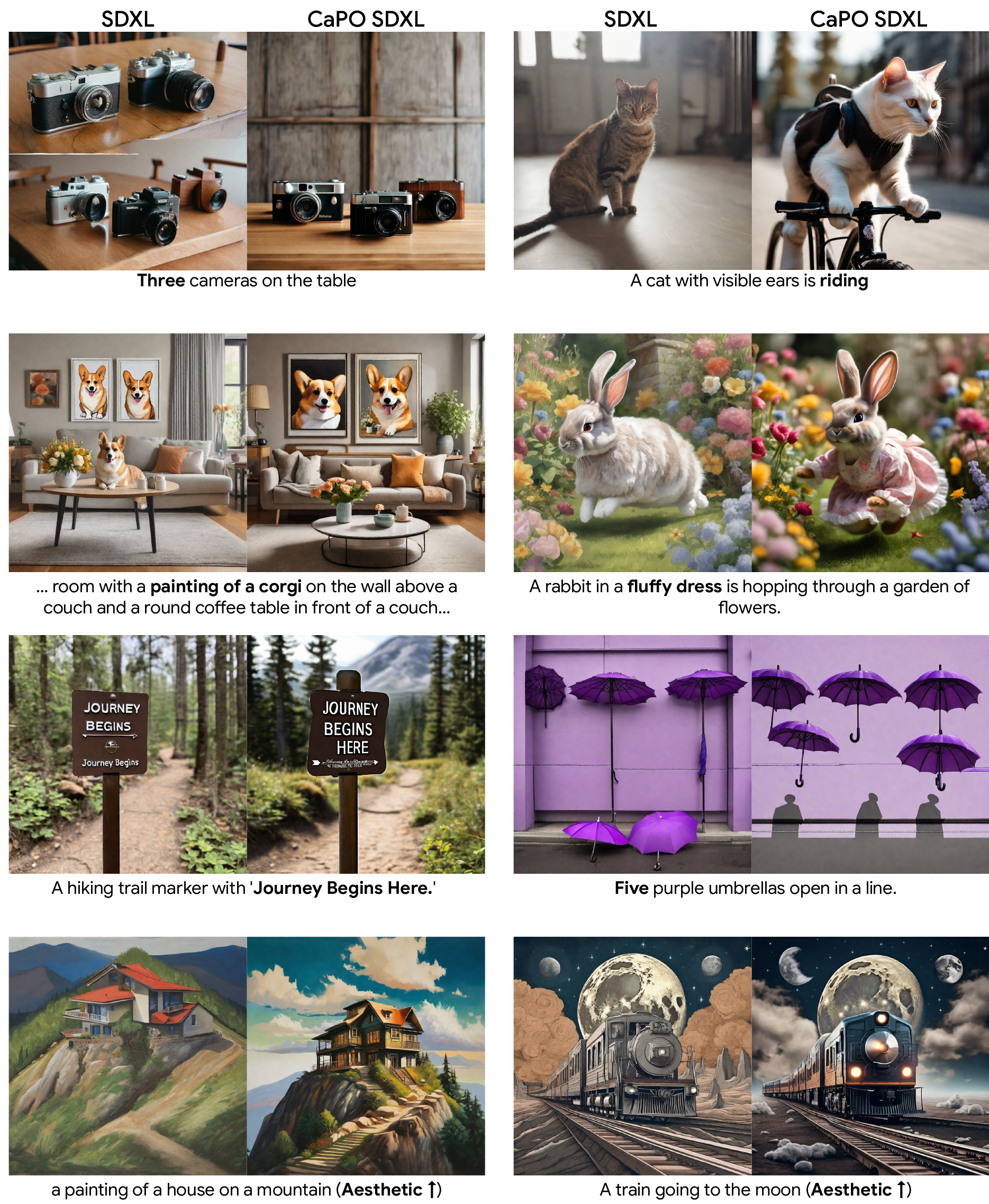}
    \vspace{-10pt}
    \caption{
    \textbf{Additional qualitative comparison between CaPO SDXL and SDXL.}
    We provide additional qualitative comparison between CaPO SDXL and SDXL. The CaPO SDXL model demonstrates better image-text alignment (\emph{e.g.}, counting, attribute binding, etc), as well as image aesthetics (\emph{e.g.}, artistic style, detail, etc). We bold the text to highlight the prompts that demonstrate improvement in image-text alignment, and ({\bf Aesthetic $\uparrow$}) to demonstrate the improvement in image aesthetic quality.
    }
    \label{fig:qual_sdxl}
    \vspace{-20pt}
\end{figure}
\newpage
\begin{figure}[!htb]
    \small\centering
    \includegraphics[width=0.98\columnwidth]{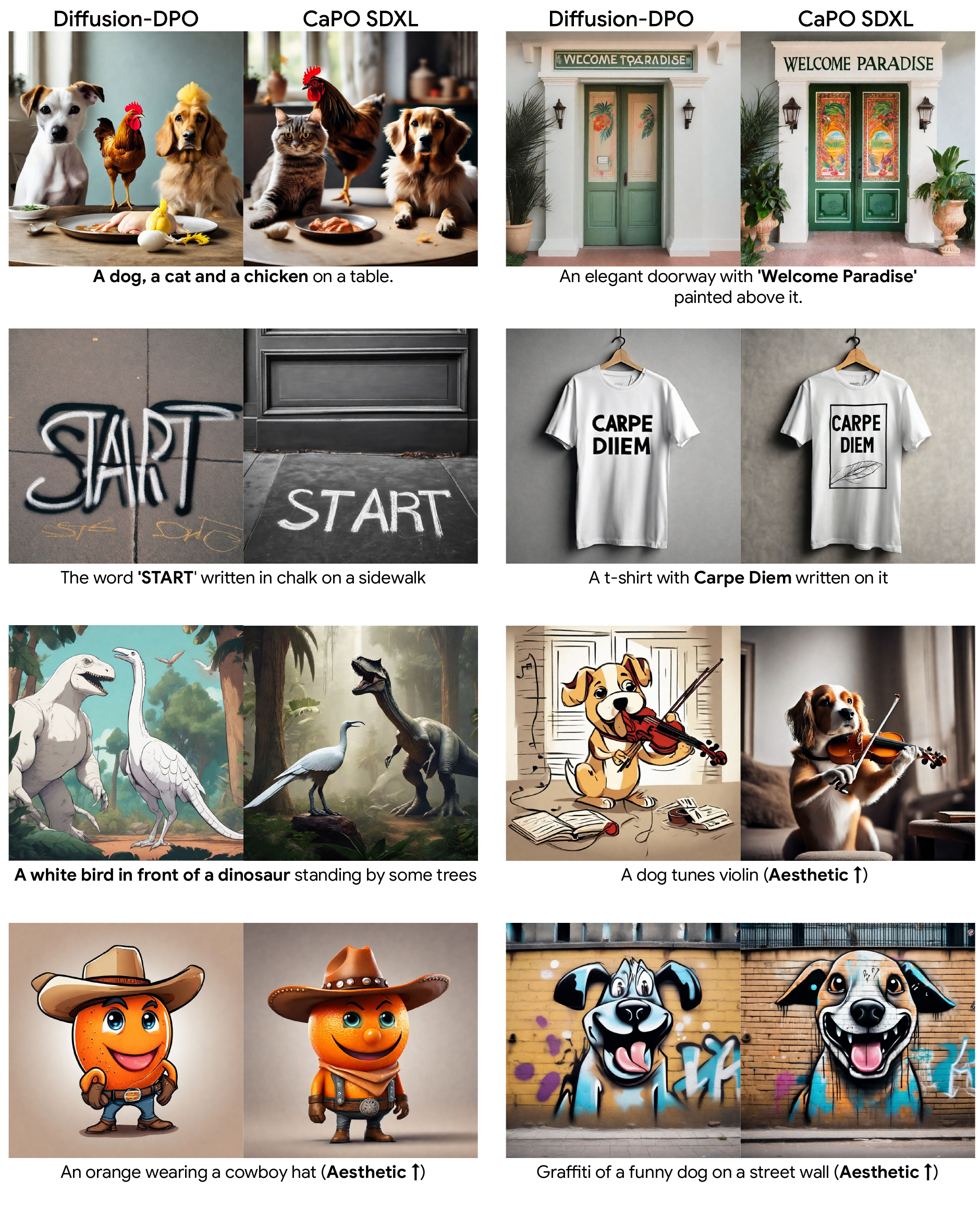}
    \vspace{-10pt}
    \caption{
    \textbf{Additional qualitative comparison between CaPO SDXL and Diffusion-DPO~\citep{wallace2023diffusion}.}
    We provide additional qualitative comparison between CaPO SDXL and Diffusion-DPO~\citep{wallace2023diffusion}. 
    CaPO SDXL shows better image-text alignment and image aesthetics compared to Diffusion-DPO without using any human annotated data.
    We bold the text to highlight the prompts that demonstrate improvement in image-text alignment, and ({\bf Aesthetic $\uparrow$}) to demonstrate the improvement in image aesthetic quality.
    }
    \label{fig:qual_sdxl_dpo}
    \vspace{-20pt}
\end{figure}
\newpage
\begin{figure}[!htb]
    \small\centering
    \includegraphics[width=0.98\columnwidth]{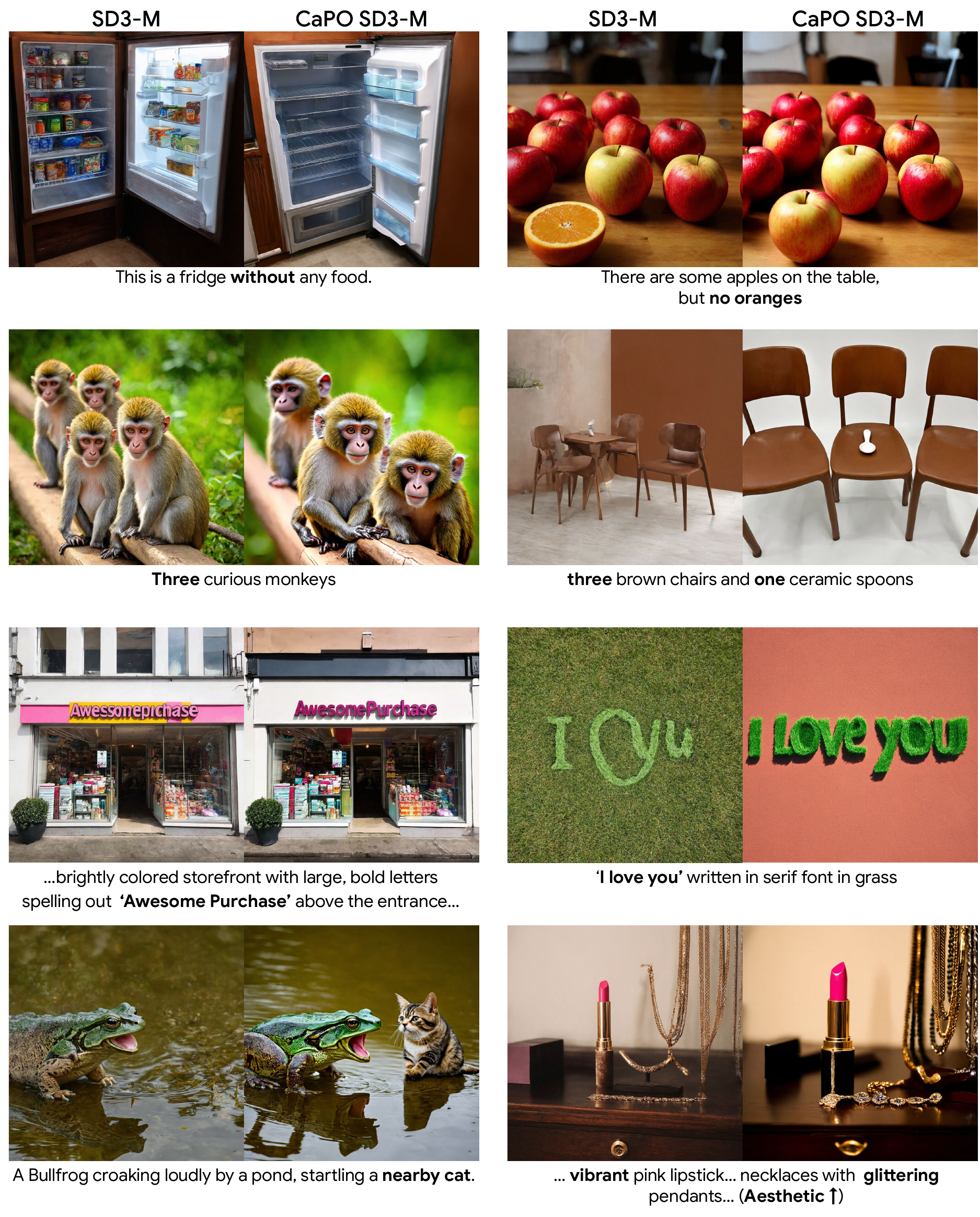}
    \vspace{-10pt}
    \caption{
    \textbf{Additional qualitative comparison between CaPO SD3-M and SD3-M.}
    We provide additional qualitative comparison between CaPO SD3-M and SD3-M. CaPO SD3-M shows better image-text alignment, \emph{e.g.}, negation (first row), counting (second row), visual text rendering (third row). Also it demonstrates better image aesthetics (fourth row right).
    We bold the text to highlight the prompts that demonstrate improvement in image-text alignment, and ({\bf Aesthetic $\uparrow$}) to demonstrate the improvement in image aesthetic quality.
    }
    \label{fig:qual_sd3m}
    \vspace{-20pt}
\end{figure}

\vspace{10pt}

\end{document}